\documentclass{article}

\usepackage{microtype}
\usepackage{graphicx}
\usepackage{subcaption}
\usepackage{booktabs}
\usepackage{hyperref}
\usepackage{multirow}
\usepackage{amsmath}
\usepackage{amssymb}
\usepackage{mathtools}
\usepackage{amsthm}
\usepackage{dblfloatfix}
\usepackage[section]{placeins}

\usepackage[preprint]{icml2026}

\usepackage[capitalize,noabbrev]{cleveref}

\theoremstyle{plain}

\theoremstyle{definition}

\theoremstyle{remark}

\icmltitlerunning{Automatically Interpreting Attribution Graphs via Probe Prompting}

\begin{document}

\twocolumn[
  \icmltitle{Automatically Interpreting Attribution Graphs via Probe Prompting}

  \begin{icmlauthorlist}
    \icmlauthor{Giuseppe Birardi}{orma}
    \icmlauthor{Gonçalo Paulo}{eai}
  \end{icmlauthorlist}

  \icmlaffiliation{orma}{Orma Lab Srl}
  \icmlaffiliation{eai}{EleutherAI}

  \icmlcorrespondingauthor{Giuseppe Birardi}{giuseppe.birardi@ormalab.it}

  \icmlkeywords{mechanistic interpretability, attribution graphs, sparse features, causal validation, steering, supernodes}

  \vskip 0.3in
]

\printAffiliationsAndNotice{}

\begin{abstract}

Even though we know the precise computations that lead from a large language model (LLM) input to its output, this computation is too complicated to interpret. We can try to make it simpler by creating a sparse computational graph that captures most of the model behavior with the smallest number of computational nodes. Cross-layer transcoders (CLT) decompose the dense computations of the MLP but the resulting circuits still contain thousands of nodes even for short prompts. Existing automated interpretation methods label individual features from corpus activations, which are often not validated by causal intervention. We introduce \emph{probe prompting}, a transparent rule-based pipeline that groups the features of an attribution graph into concept-aligned supernodes from their responses on a small set of concept-targeted probe prompts, summarized as Cross-Prompt Activation Signatures (CPAS). Across four factual domains, on Gemma-2-2B with a public CLT dictionary and $44{,}596$ entity-swap interventions, we find that the labeled supernodes beat random and influence matched baselines.
Code, datasets, and an interactive demo are released anonymously as a reusable harness for calibrating supernode labels against causal interventions.
\end{abstract}

\section{Introduction}
\label{sec:intro}

A part of mechanistic interpretability research in large language models (LLMs) is focused in decomposing the representations used by these models and on simplifying the computational graph in such a way as a to make it human understandable, so called reverse engineering. One way researchers have done this decomposition is by training sparse autoencoders (SAEs) and their variants~\cite{bricken2023monosemanticity,templeton2024scaling,crosscoders2024,marks2024dictionary}. SAEs are trained with the objective of reconstructing model activations, be them the residual stream, the output of MLP or even of attention heads, while using a sparsely activating basis (features) which is hopefully more interpretable.

Automated interpretation of individual features by LLM scoring~\cite{bills2023autointerp,paulo2024autointerp,templeton2024scaling} produces per-feature labels from corpus activations but does not yield circuit-level interaction. Attribution graphs~\cite{ameisen2025circuittracing,lindsey2025biology} trace the influence of features through the computational graph, but most graphs still have too many nodes to manually interpret. A single Gemma-2-2B graph for a short factual prompt routinely contains 600--5{,}000 cross-layer transcoder (CLT) features and edges between them~\cite{ameisen2025circuittracing,lindsey2025biology}, and manual reading by an experienced circuit tracer has been reported to take on the order of two hours per prompt~\cite{neuronpedia2025podcast}. As mechanistic interpretability moves from individual case studies to large-scale catalogs of circuits~\cite{hannaameisen2026planning,marks2024dictionary}, automated circuit-level grouping tied to behavior is required.

We propose \emph{probe prompting}, a transparent rule-based pipeline that converts an attribution graph into a compact set of concept-aligned supernodes. For each candidate feature in the graph we run a small set of concept-targeted probe prompts and summarize the feature's responses as a Cross-Prompt Activation Signature (CPAS). Deterministic threshold rules map each signature to one of four functional roles, and features that share role and name are merged into a supernode. Every assignment is traceable to a specific threshold crossing, so a researcher can audit and edit the grouping. We can then produce controlled intervention experiments that can evaluate how good the supernode labels are.

\begin{figure*}[t]
  \centering
  \includegraphics[width=0.95\textwidth]{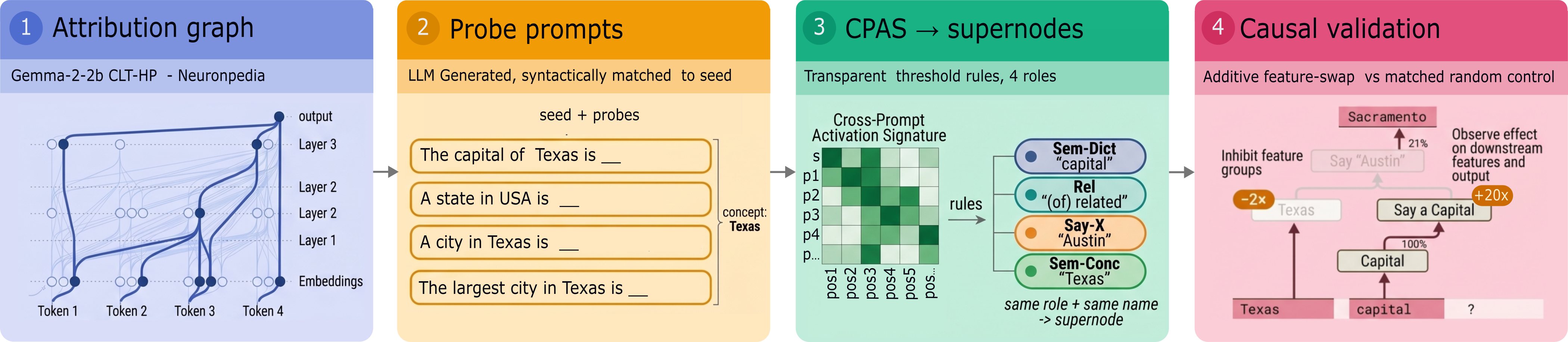}
  \caption{Pipeline overview. Four stages: (1) attribution graph generation (through Neuronpedia); (2) probe prompts (LLM-generated, syntactically matched to seed); (3) Cross-Prompt Activation Signatures followed by transparent rule-based supernode typing; (4) causal validation via additive feature-swap interventions with structurally matched random controls (same feature count, same per-layer histogram, sampled outside concept-matching supernodes). The unit of analysis is the supernode, validated operationally by the Specificity and Matched-Control conditions of \S\ref{sec:results:primary}.}
  \label{fig:pipeline}
\end{figure*}

The paper contributes:
\begin{itemize}\itemsep0pt
  \item A transparent rule-based pipeline that groups the features of an attribution graph into concept-aligned supernodes from Cross-Prompt Activation Signatures (Section~\ref{sec:method}).
  \item A fitness check for supernode labels, with a matched-random-control swap protocol that is grouping-method-agnostic and applicable to any future grouping, including learned ones (Sections~\ref{sec:method}), as well as a influence matched protocol that swaps the most influential features.
  \item Evidence that the proposed protocol labeled supernodes pass this test in four different factual-recall domains, showing the generalizability of the technique (Section~\ref{sec:results}).
  \item A open release of a $44{,}596$-run intervention dataset, anonymous code, and an interactive demo (Appendix~\ref{sec:repro}).
\end{itemize}

\section{Related Work}
\label{sec:related}

\paragraph{Attribution graphs and replacement models.}
Attribution graphs operationalize feature$\to$logit pathways in a local replacement model that linearizes the residual stream through CLT or SAE features and freezes attention~\cite{ameisen2025circuittracing,lindsey2025biology}.  \href{Neuronpedia}{https://www.neuronpedia.org/} exposes these graphs for selected models and displays per-feature cards at scale~\cite{neuronpedia2025,lindsey2025landscape}. \texttt{Circuit-tracer} provides the open-source reference implementation~\cite{circuittracer2025} for computing these graphs. Our work operates entirely \emph{downstream} of attribution: we do not propose a new attribution method, but we do produce supernodes that Neuronpedia tooling can render and pin. Our work proposes a clustering technique that reduces the human labor required to produce supernode labels.

\paragraph{Causal circuits with transcoders.}
\citet{hannaameisen2026planning} use cross-layer transcoder feature circuits to study whether language models latently plan rhyming words, structuring their analysis around two falsifiability conditions (predict-token-before-emission, causally-affect-token). \citet{marks2024dictionary} apply sparse-feature circuit analysis to subject--verb agreement; we contribute a domain-general harness and a proposal on how to generate groupings of attribution graphs on scale.

\paragraph{Causal interpretability and matched random controls.}
\citet{geiger2025causal} formalize three levels of interpretive claim and emphasize that \emph{distinguishing} levels requires intervention. \citet{heap2025sparse} show that SAE auto-interpretability scores~\cite{bills2023autointerp,paulo2024autointerp} and standard SAE reconstruction metrics can be similar for randomly-initialized and trained transformers, highlighting the need for metrics anchored to causal effect. Our matched-random-control protocol (\S\ref{sec:method:harness}) is designed to address this critique: any structural property a random feature set shares with the labeled one (count, layer histogram, eligibility) is held constant; only the concept alignment changes. We also compare our intervention technique with substituting the most influential features, showing that our grouping is working not only by increasing the number of features steered.

\section{Method}
\label{sec:method}

\subsection{Domains and seed prompts}
\label{sec:method:domains}
We set  four domains spanning two-hop factual recall in different relational structures and answer templates. Each domain has a fixed seed-prompt with three semantic fields (input, intermediate, answer): 
\begin{itemize}\itemsep0pt
\item USA: ``capital of state containing $X$''; 50 entities, 2{,}450 non-identity swap pairs.
\item Books: ``character $X$ appears in a book by''; 10 entities, 90 non-identity swap pairs.
\item Products: ``founder of the company that makes $X$''; 12 entities, 132 non-identity swap pairs.
\item Paintings: ``first name of the painter of $X$''; 10 entities, 90 non-identity swap pairs.
\end{itemize}

\subsection{Setting and attribution-graph generation}
\label{sec:method:setting}
We use Gemma-2-2B with the public CLT dictionary ($\sim$2.5M features over 26 layers; \citealp{hanna2025clthp,ameisen2025circuittracing}). Attribution graphs are generated via the Neuronpedia API; exact API parameter names and values are in \cref{appx:pipeline}. The replacement model freezes attention during graph computation.

\begin{figure*}[!t]
  \centering
  \includegraphics[width=\textwidth]{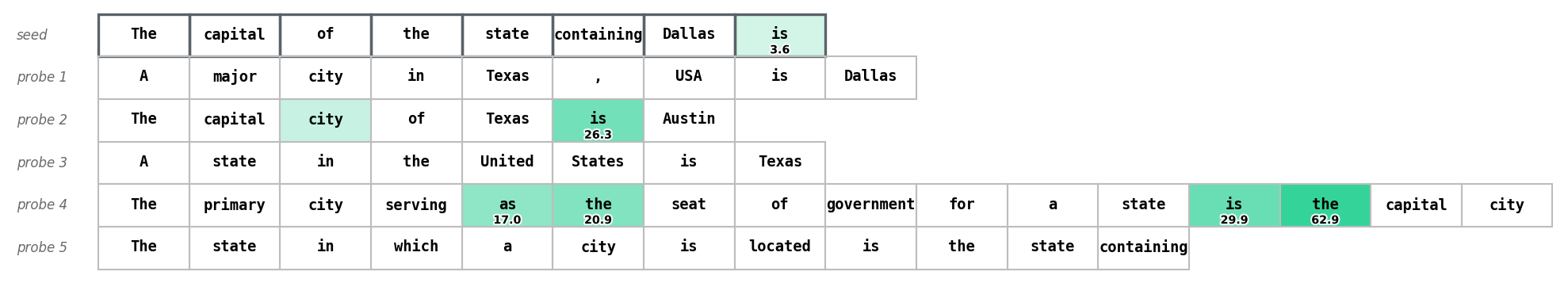}
  \caption{Feature \texttt{20-clt-hp:74108} (Say ``capital''). Activations on the seed prompt and five concept-targeted probes; peak intensity is shaded in green. On the seed alone the feature looks like a peak on the copula \texttt{is}. Across probes it instead peaks on a variety of functional tokens (\texttt{is}, \texttt{the}, \texttt{as}) whose only consistent property is that they sit immediately before the semantic target \emph{capital}; the maximum activation ($63.31$) lands on the \texttt{the} preceding ``capital'' in probe~4. The cross-prompt pattern identifies this feature as a procedural output-promotion (\textsc{Say-X}) feature for \emph{capital}, not a semantic detector for \texttt{is}.}
  \label{fig:cpas-say-capital}
\end{figure*}

\subsection{Probe prompt generation}
\label{sec:method:probes}
Given a prompt used to generate the attribution graph $p$ (e.g., ``the capital of the state containing Dallas is''), we use an LLM to generate a small set of concept-targeted probes. First, we extract the $K$ most salient concepts from $p$ (and, when available, from the model's output), each tagged with a short description. The second pass turns the extracted concepts into probe prompts that share the seed's surface structure (same prepositions, same answer position, comparable length) while varying the content entities. A typical seed yields 4--5 probes. The full system prompt and the concept-to-probe template are reproduced in \cref{appx:pipeline}.

\subsection{Cross-Prompt Activation Signatures (CPAS)}
\label{sec:method:cpas}
A peak on a single prompt is ambiguous: a feature that activates on \texttt{is} in the seed could be a dictionary detector for the copula, a context-specific feature, or a procedural feature that promotes whatever output the model is about to emit. The CPAS resolves the ambiguity by aggregating peaks across deliberately varied probes (\cref{fig:cpas-say-capital}).

For each (feature, probe prompt) pair we record a small set of per-probe measures: at each token position we measure the feature's peak activation, how sparsely the feature fires across the probe set, and a robust $z$-score and cosine similarity of the feature's activations relative to the seed prompt. Aggregating across probes yields a compact per-feature signature (the CPAS): a handful of summary numbers capturing how consistent the peak token is across probes, how many distinct tokens the feature peaks on, and the feature's confidence in its dominant role. Formal definitions and raw-measure specifications are in \cref{appx:pipeline}.

\subsection{Functional vs.\ semantic tokens; target-token mapping}
\label{sec:method:funcsem}
Tokens are labeled \emph{functional} (eg. copulas, articles, prepositions) or \emph{semantic} (content-bearing). When a feature peaks on a functional token, a $\pm 7$-token directional search identifies the nearest semantic peak (e.g., forward for articles; backward for the possessive \texttt{’s} ). 

\subsection{Four functional roles}
\label{sec:method:rules}
Each feature is assigned one of four functional roles. The four-role vocabulary is an empirical extension of the supernode types that appear in circuit-tracing case studies~\cite{lindsey2025biology,ameisen2025circuittracing}; we found that these four roles covered the vast majority of features we encountered during pilot analyses, and we formalize them here as the pipeline's output types.

\begin{itemize}\itemsep0pt
\item \textbf{Semantic-Dictionary (\textsc{Sem-Dict}).} The feature fires consistently on the \emph{same} semantic token across probes---a dictionary-like detector for a specific concept (e.g., a feature that reliably peaks on ``Texas'' whenever Texas is mentioned).
\item \textbf{Semantic-Concept (\textsc{Sem-Conc}).} The feature peaks on semantic tokens across a small family of related tokens rather than on a single one---a concept-level rather than token-level detector, typical of middle layers.
\item \textbf{Relationship (\textsc{Rel}).} The feature does not concentrate its activity on any particular token; it fires diffusely across the probe, with a comparatively dense activation pattern. These features appear to encode a relation or context rather than a named entity.
\item \textbf{Say-X (\textsc{Say-X}).} The feature peaks on functional tokens (e.g., \texttt{is}, \texttt{the}) in predictable positions relative to a semantic target, and sits in the later half of the network. After the $\pm 7$-token directional search (\cref{sec:method:funcsem}), the feature is named by the \emph{target} semantic token that follows or precedes the functional peak.
\end{itemize}

The exact thresholds and priority ordering are in \cref{appx:pipeline}. A post-hoc clustering analysis of the per-feature CPAS metrics confirms that the four-role partition is best read as a deliberate \emph{coarse-graining} of a finer natural geometry; details in \cref{appx:clustering}.

\subsection{Supernode formation and naming}
\label{sec:method:supernodes}
Same-role same-name features form a supernode. Naming is role-specific.

\emph{Semantic} features (both \textsc{Sem-Dict} and \textsc{Sem-Conc}) are named by the semantic token at which they activate most strongly: a feature that consistently peaks on ``Texas'' receives the name \emph{Texas}.

\emph{Say-X} features are named by the semantic token they appear to promote: e.g., \emph{Say (Austin)} for a feature peaking on \texttt{is} immediately before ``Austin''.

\emph{Relationship} features are named by the highest-activation semantic token . The format is \emph{(token) related}, producing names like \emph{(containing) related} for features that fire diffusely on spatial-relationship phrases.


\subsection{Causal validation of grouping}
\label{sec:method:harness}
We test whether supernodes have an operational causal consequence. For a source--target pair \$(e\_A, e\_B)\$, the intervention asks: if we suppress the features labeled as related to the source entity and amplify the features labeled as related to the target entity, does the model redirect its answer from \$e\_A\$ toward \$e\_B\$? 
For each intervened feature, $M \cdot v_{\mathrm{orig}}$ scales the decoder vectors and adds the result to the residual stream at the downstream layers. Attention is not frozen during the intervention; positive results are stronger evidence than in patching of frozen-attention where some direct effects are forced by construction~\cite{ameisen2025circuittracing}.

We use two primary metrics: \textbf{Hit\%}, the fraction of pairs whose generated continuation contains the target's first-subword token; \textbf{vsMax}, the maximum over the generated trajectory of the target's logit minus the best other answer in the domain (positive means the target is ahead). A small set of secondary diagnostics is described in \cref{appx:pipeline}.

For each swap pair we construct a random feature set with the same feature count and per-layer histogram as the labeled intervention, drawing features from \emph{outside} every concept-aligned supernode in the domain so that the only property that varies is concept alignment. Three deterministic replicates per pair yield Hit-rate and vsMax distributions over random feature sets. 

As a second baseline, we also run an influence-matched top-$K$ control. For each labeled swap, this baseline selects the smallest prefix of features ranked by graph node influence whose cumulative source- and target-side influence matches the labeled intervention budget. The top-$K$ control uses the same ablation/amplification signs, but discards CPAS labels, field names, and supernode membership. It therefore tests whether the effect is explained merely by intervening on high-influence graph nodes, rather than by concept-aligned grouping; details are in \cref{appx:topkim}.

Each domain has 3 semantic fields connected to the two-hop logic (input, intermediate, answer). For each pair, 7 variants are run: 3 single-field, 3 two-field, 1 all-three. Each selected field drives both ablation and amplification; this isolates which fields carry the causal signal.

When a swap misses at the default $M_{\mathrm{amplify}}{=}20$, we search for a better amplifier in two phases. Phase~1 probes a coarse geometric grid and stops at the first hit. If Phase~1 finds no hit, Phase~2 uses the fact that the steering effect typically has a sharp onset: we compute the KL divergence of the steered output against the unsteered baseline at each probed $M$, locate the interval in which KL rises most steeply, and binary-refine inside that interval for 6 further steps, accepting a hit at any refinement point. Full pseudocode in \cref{appx:msearch}.

\begin{figure}[t]
  \centering
  \includegraphics[width=\linewidth]{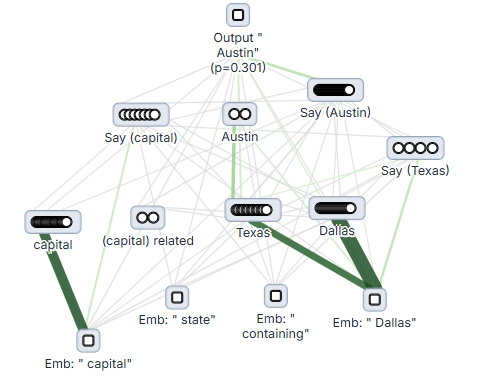}
  \caption{Concept-aligned subgraph for the seed ``the capital of the state containing Dallas is''. Token embeddings (\texttt{Emb}) are at the bottom; the model's predicted continuation (\emph{Austin}, $p{=}0.301$) is at the top. Each supernode (rounded box) groups same-role same-name features and is named by its role: semantic (\emph{capital}, \emph{Texas}, \emph{Dallas}, \emph{Austin}), relationship (\emph{(capital) related}), or output-promoter (\emph{Say (capital)}, \emph{Say (Texas)}, \emph{Say (Austin)}). Edge thickness is proportional to influence on the output.}
  \label{fig:circuit-texas}
\end{figure}
\section{Results}
\label{sec:results}
\subsection{State capital case study}
\label{sec:results:primary}

\begin{table*}
\centering
\small
\caption{\textbf{Prompt probes beat matched-random and influence-matched top-$K$ across 4 domains.} 
Each row reports three methods on the same per-domain entity set. 
\emph{Ours}: per-pair best field-additivity variant with adaptive $M$-search on labeled features (\cref{appx:msearch}, \cref{tab:msearch-rescue}). 
\emph{Rand.}: matched-random control under the same per-pair best-of construction (the best across the $3$ replicates with adaptive $M$-search; \cref{appx:specificity}, \cref{tab:fullscale-spec}). 
\emph{Top-$K$}: per-pair influence-matched top-$K$-by-graph-influence baseline  with adaptive M -search (\cref{appx:topkim}, \cref{tab:topk-im}). 
\emph{Hit\%}, the fraction of pairs
whose generated continuation contains the target’s first-subword token.
\emph{vsMax} is the mean per-pair best logit margin against the strongest competing dataset answer}
\label{tab:headline}
\begin{tabular}{lr@{\hspace{4pt}}rr@{\hspace{4pt}}rr@{\hspace{4pt}}rrl}
\toprule
& & \multicolumn{2}{c}{Ours} & \multicolumn{2}{c}{Rand.} & \multicolumn{2}{c}{Top-$K$} & \\
\cmidrule(lr){3-4}\cmidrule(lr){5-6}\cmidrule(lr){7-8}
Domain & $N$ & Hit\% & vsMax & Hit\% & vsMax & Hit\% & vsMax\\
\midrule
USA       & 2{,}450 & 72.8 & $+6.15$ &  0.7 & $-1.23$ & 4.2 & $-1.92$  \\
Books     &     90  & 77.8 & $+10.38$&  0.0 & $+0.19$ & 4.4 & $+1.57$  \\
Products  &    132  & 41.7 & $+5.42$ &  7.6 & $+1.17$ & 1.1 & $+0.63$  \\
Paintings &     90  & 18.9 & $+3.45$ &  1.1 & $+1.27$ & 7.1 & $+0.16$  \\
\bottomrule
\end{tabular}
\end{table*}
\citet{lindsey2025biology} use attribution graphs to investigate two-step reasoning with the prompt ``The capital of the state containing Dallas is'', and we start from the same template. The Gemma~2 CLT attribution graph for this prompt contains $1{,}182$ features at our cumulative-influence threshold. Running these features through the probe-prompt pipeline (\cref{fig:pipeline}) places $458$ of them into eight concept-aligned supernodes  (\emph{capital}, \emph{state}, \emph{Texas}, \emph{Dallas}, \emph{Austin}, \emph{Say (capital)}, \emph{Say (Texas)}, \emph{Say (Austin)}); the resulting subgraph keeps a Neuronpedia completeness of $0.83$ and a replacement of $0.53$.  \emph{Completeness} is the fraction of incoming edges to all nodes of the subgraph that originate from grouped features weighted by influence on the output, while  \emph{Replacement} is the fraction of end-to-end influence from input tokens to output logits that flows through grouped features.   As a reference point, the human-annotated subgraph featured on Neuronpedia for the same prompt\footnote{Public graph slug \texttt{gemma-fact-dallas-austin} on \url{neuronpedia.org}.} pins $21$ features into five named supernodes and reaches a completeness of $0.70$ and a replacement of $0.30$.

\begin{figure}[!t]
  \centering
  \includegraphics[width=\linewidth]{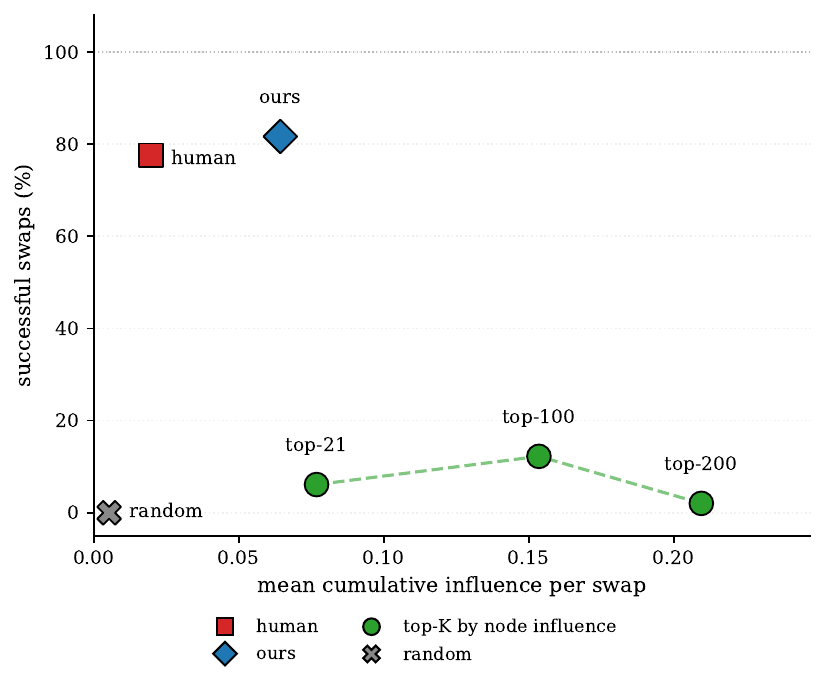}
  \caption{Successful swaps (\%) vs. mean cumulative influence per swap. Each point denotes one approach evaluated on the $49$ non-Dallas USA states. The target is fixed to Dallas/Texas/Austin, since human annotation is available only for that graph. The y-axis reports the successful-swap rate: the fraction of source states for which at least one field variant redirects the model's predicted capital to \emph{Austin}. The x-axis reports the mean, across swaps, of the cumulative $\mathrm{node\_influence}$ carried by the amplified features in the target graph. Both labeled methods reach approximately $80\%$ successful swaps, whereas the top-$K$ control family saturates near $10\%$ despite using $1.09\times$--$2.93\times$ more per-swap influence than our method.}
  \label{fig:topk-saturation-full50}
\end{figure}

We measure how useful the automated clustering is by entity-swap experiments: for a (source, target) pair we ablate features from the source state and amplify the corresponding features from the target state, and check whether the model's predicted capital is redirected to the target's capital. Because the only human-labeled graph available is the Dallas one, every swap uses Dallas as target and one of the other 49 states as source; the target supernodes come from the human annotation, while the source supernodes always come from the auto pipeline (no human annotation exists for the other 49 graphs). Across the $49$ sources, our supernodes redirect the prediction to the target capital on $40/49$ sources versus $38/49$ for the human-labeled supernodes and $0/49$ for the random control; the top-K node-influence baseline saturates at $6/49$  when given $\sim$100 features per call (\cref{fig:topk-saturation-full50}). 

\begin{figure}[!t]
  \centering
  \includegraphics[width=\linewidth]{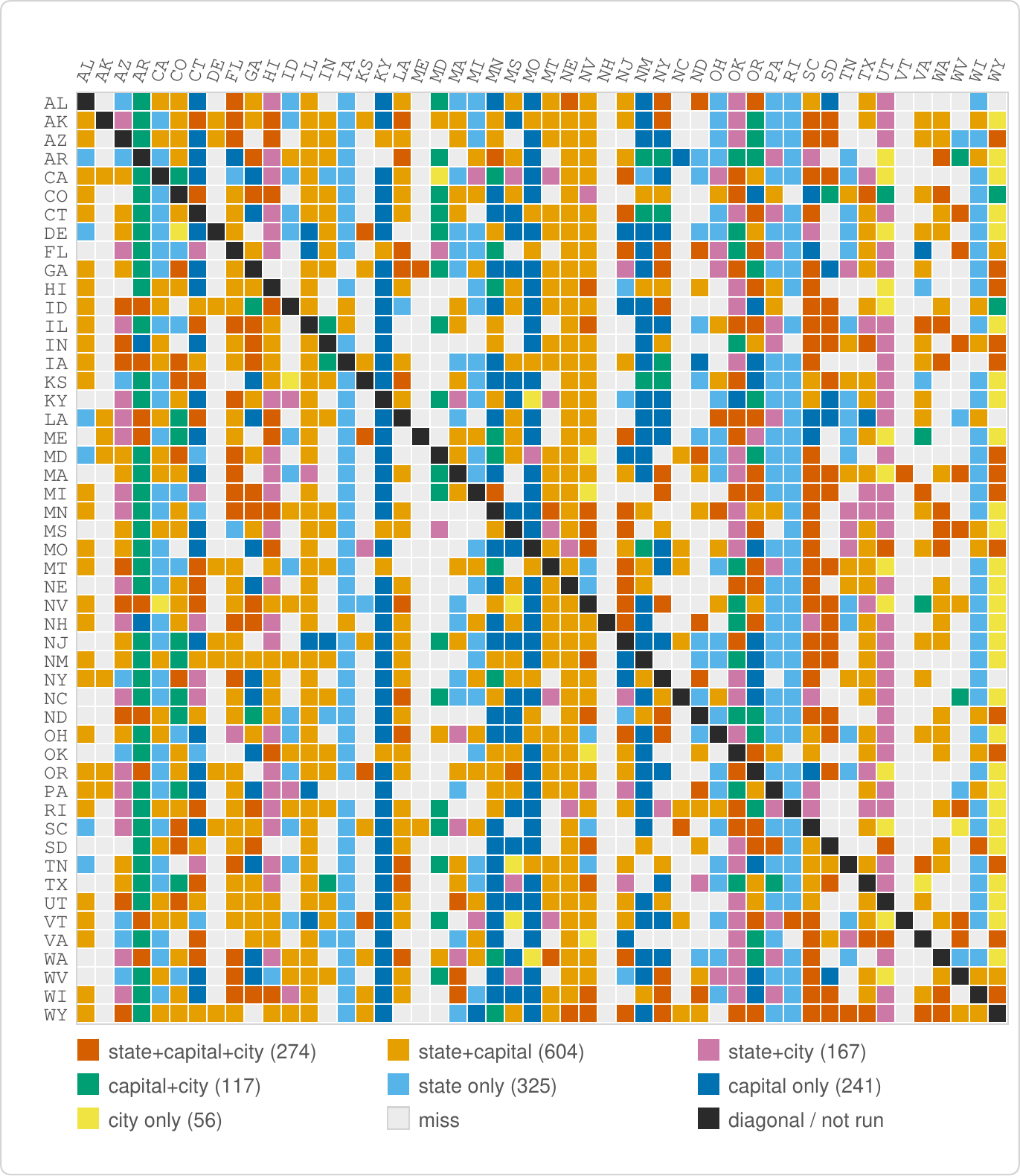}
  \caption{Source $\to$ target swap matrix on the 50-state USA panel
    ($n{=}2{,}450$ pairs, hit-rate $72.8$\%). By default we steer on all
    three concept fields (state + capital + city); each cell additionally
    considers every smaller field subset and reports the winning
    configuration. The winner is the variant that hits the target capital
    (steered top-1 = target capital first-token); when several variants
    hit, we prefer the one that maximises the steered margin of the target
    logit over the next-best capital in the dataset.
    Cells are coloured by the winning field subset; the 3-field default
    wins only $15$\% of cells, the rest succeed with a strict subset. Grey cells: miss; black on
    the diagonal: pair not run.}
  \label{fig:swap-matrix}
\end{figure}

We find that it is necessary to steer different sets of clusters to correctly induce the entity swap - one can either steer the 'input' field, the city name, the 'intermediate' field, the state, or the 'answer' field, the correct capital, as well as any combination. When performing all possible state swaps, the predicted capital is redirected to the intended target on $72.8\%$ of pairs, see \Cref{fig:swap-matrix}. By default we steer on the 3 concepts but if another steering combination is better we color code the cell by the combination used. The vertical striping in the swap matrix further suggests that the optimal subset is largely target-conditioned: for a given target state, the same field combination often transfers to many different source states.While most states can be used as sources, some of the states can barely be used as targets, like NV, ME, NH, VT AK. This likely reflects the fact that their capitals already have low baseline logit probability and are often not top-logit predictions even in the unsteered setting.

\subsection{Generalization to other domains}

When developing the probing technique, we focused on the state capital prompt template, which might 'overfit' our decisions to making supernodes that work on this specific setting but don't translate well to other types of prompts and problems. Because of this we create other 3 two-hop datasets, and use our pipeline as is to produce labeled supernodes for graphs computed for these tasks. 

We find that in all domains, our labeled supernodes lead to both a higher number of entity swaps (Hit\%) as well as a higher average larger logit gap between the target token and other valid answers (Table~\ref{tab:headline}), when compared with randomly selecting features, as well as using top-$k$ features by graph influence at the same per-pair influence budget (\cref{appx:topkim}). On the other hand, while both the States and the Book datasets have a majority rate of swapping entities, the other datasets don't have such high steering performances. Because there are no human labeled equivalents, it is hard to establish a good base performance, being unclear if the failure comes from our pipeline, the attribution graphs, or the CLT figures.

\begin{figure}[!t]
  \centering
  \includegraphics[width=1\linewidth]{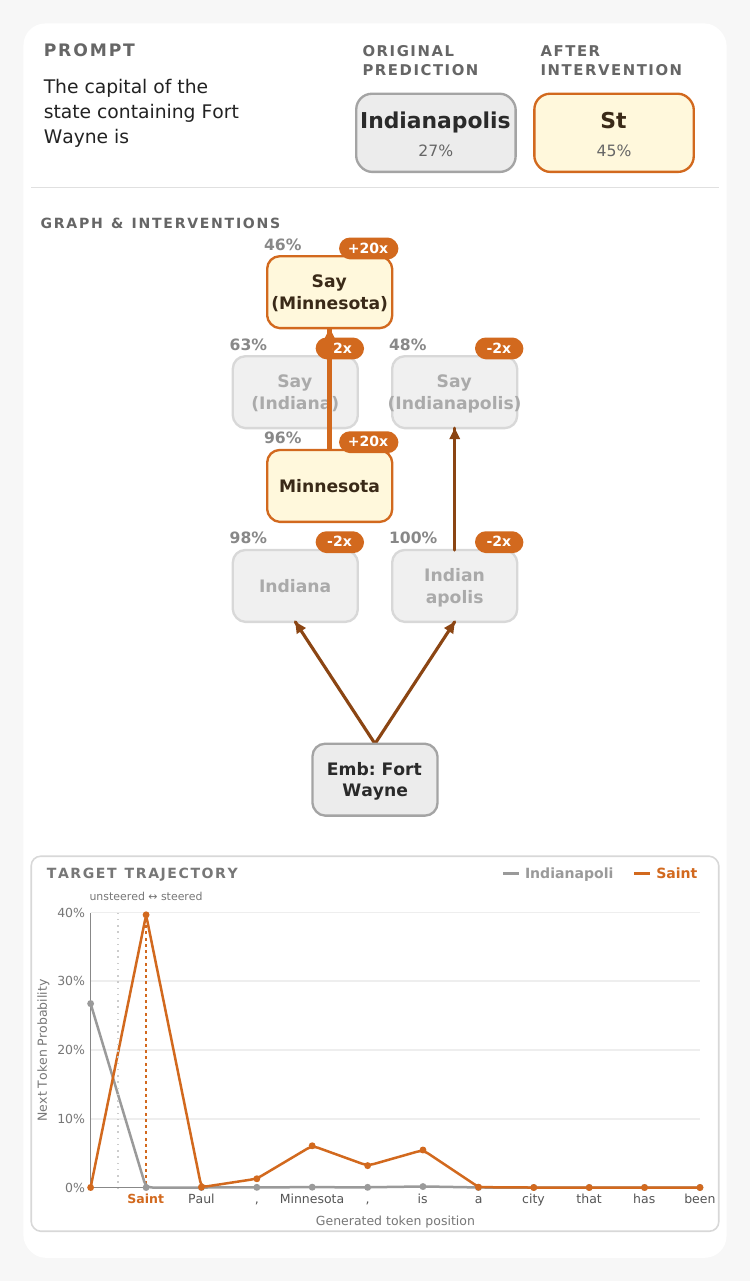}
  \caption{A representative entity-swap intervention on the USA panel:
    Indiana$\to$Minnesota, probed via ``the state containing Fort Wayne''.
    The top row shows the prompt and the pre/post-intervention prediction
    (\texttt{Indianapolis} $\to$ \texttt{St}, i.e.\ \texttt{Saint~Paul}). The
    middle row shows the labeled supernodes with ablation ($-2\times$) of the
    source state/capital nodes and amplification ($+20\times$) of the target
    state/capital nodes. The bottom row shows the next-token probability sweep
    at $M{=}0$ vs.\ $M{=}20$, with the target trajectory (\texttt{Saint}) rising
    at the steered position while the original (\texttt{Indianapolis}) collapses.}
  \label{fig:swap-examples}
\end{figure}

Also in these domains we find that correctly choosing the subset to steer can drastically change the effectivness of steering, with the most effective concepts to steer being the intermediate and answer fields, working better than steering all three fields. (\cref{appx:fieldadd}).

We find that a single steering strength for all the experiments is not the most optimal and that there are two different steering strengths, with peaks near $M{\sim}2.4$ ($47$\% of labeled hits) and $M{\sim}6.9$ ($40$\%), that recover most of the swap performance (\cref{appx:msearch}). Values outside these regimes can fail, with low $M$ leaving the source attractor intact and high $M$ sometimes overshooting the target and disrupting the continuation. These is different from what was reported in the single-feature SAE setting of \citet{templeton2024scaling}, in which clamping at roughly $\pm 10\times$ maximum observed activation saturates the behavior.

\section{Limitations}
\label{sec:limitations}

\paragraph{No grouping-method baseline.}
We compare labeled supernodes against matched random controls rather than against geometric or learned clustering because there is no canonical feature-clustering baseline in the SAE/CLT literature. The harness (\S\ref{sec:method:harness}) is method-agnostic and we encourage future groupings to be evaluated on it.

\paragraph{Attention-circuit blindspot.}
Attribution graphs freeze attention; our swaps run with attention free. The asymmetry plausibly accounts for part of the residual miss rate; attention-aware variants are future work (\cref{sec:discussion}).

\paragraph{Threshold sensitivity is unmeasured.}
Decision-rule thresholds were iteratively refined on pilot circuits; we did not run a formal $\pm 10$--$20\%$ sensitivity sweep. Thresholds are configurable defaults in the released code. 



\paragraph{Domain scope: two-hop factual recall only.}
Our four evaluation domains are all two-hop factual-recall prompts. Whether the pipeline transfers to relational tasks with different structure (multi-hop reasoning, antonymy, code, dialogue, instruction-following) is untested.

\paragraph{Single model and language.}
In the three-level framing of \citet{geiger2025causal} we claim Levels~1 and~2 only; we don't think that our current pipeline allows to have a Level~3 (full mechanistic identification of latent variables) understanding. All experiments use Gemma-2-2B-it with CLT-HP on English text(\cref{sec:method:domains}); cross-architecture, cross-dictionary, and multilingual generalization are open.

\section{Discussion and Conclusion}
\label{sec:discussion}

\paragraph{Methodological lessons.}
Three findings in this paper may inform intervention practice in related settings. First, matched random controls need to be \emph{structurally} matched---same feature count, same per-layer histogram, and sampled outside every concept-aligned supernode---because otherwise the labeled--random gap is easy to misstate: in three of our four domains, matched random feature sets produce \emph{more} source-suppression than the labeled intervention does, even though they redirect the model nowhere in particular. Second, the $M{=}20$ amplification default that is standard for single-feature steering does not carry over to the multi-feature CLT setting studied here: our winning values of $M$ are bimodal at roughly $2.4$ and $6.9$, and a coarse sweep over $M$ rescues a substantial fraction of misses. Third, semantic fields are not interchangeable for intervention purposes: restricting to intermediate- and answer-field features gains $14$--$33$~pp of Hit-rate over using all three fields.


\paragraph{Polysemanticity and cross-prompt activation.}
CLT and SAE features are not guaranteed to be monosemantic in practice. \citet{templeton2024scaling} document \emph{feature splitting}---a coarse feature in a smaller dictionary splits into several finer features as dictionary capacity grows---indicating that ``one feature'' is often a bundle of related meanings. \citet{lindsey2025biology} note this explicitly: some member features are polysemantic in a dictionary-wide sense, and are kept in the supernode only on account of the facet relevant to the traced prompt. \citet{balcells2024evolution} document related pass-through and specialization phenomena layer-by-layer. Probe prompting formalizes this ad-hoc practice. By comparing a feature's activation across concept-targeted probes, CPAS assigns each feature a \emph{prompt-contextual} role: the feature is monosemantic enough, in the context of the prompt family under study, to serve as the unit of a supernode, even when it is polysemantic across the dictionary as a whole. The same logic suggests that the method is not strictly tied to a trained sparse dictionary and can in principle be applied to raw MLP neurons, which are known to be substantially more polysemantic: when neurons are inspected on a small set of carefully chosen probes instead of on corpus-wide activations, a usable functional signature can emerge even without the sparsity prior.

\paragraph{Future work.}
Several directions seem natural. An obvious first extension is to make the intervention attention-aware, either by incorporating edge-attribution information or by also swapping attention-head outputs along with CLT feature activations; replicating the main findings on Gemma-2-9B, Llama-3, and alternative feature dictionaries would test how much of the current picture is specific to Gemma-2-2B-it and CLTs. A third direction is multilingual extension: our current prompt templates are English-only, and we have observed (\cref{appx:funcvocab}) that the functional-token vocabulary and the directional search it drives do not transfer cleanly to, for example, French. 

Two further directions deserve specific mention. The first concerns the \emph{probe set} itself. Our current probes are positive variations on the seed prompt (``the $\langle\mathrm{entity}\rangle$ is''), and we treat the set as fixed across features. There is no reason to do so. Contrastive or near-miss probes---probes that differ minimally from the seed along a concept dimension the model might confuse---should be considerably more discriminating. At longer context lengths, standard for frontier models, one can imagine agentic variants of the harness that treat probe design as a detective game: an agent with a bounded action budget chooses which substring or position to intervene on next, based on what the previous probe revealed. The second direction concerns the feature-role taxonomy itself. The four-role vocabulary (Semantic-Dictionary, Semantic-Concept, Relationship, Say-X) is an empirical extension of the roles that appear in circuit-tracing case studies~\cite{lindsey2025biology}, and the fact that it generalizes across four unrelated domains suggests there is a useful ``model-biology'' object hiding behind it; sharpening this taxonomy---adding roles, subdividing existing ones, or reducing them---is a research question in its own right. Finally, the grouping machinery itself is open: learned CPAS classifiers trained on the rule-based decisions could replace the hand-authored decision tree with a smoother map from signatures to roles.

\paragraph{Conclusion.}
Probe prompting shows that circuit-level interpretability of attribution graphs can be partially automated with a transparent, rule-based pipeline, and that the resulting labels can be used to perform entity swaps, an intervention technique that can be used to quickly evaluate whether supernode labels correctly cluster features into their actual behavior in the computational graph. 


\section{Contributions}

Giuseppe Birardi conceptualized the work, performed the experiments, analyzed the data, created the demos and wrote the initial draft. Gonçalo Paulo helped with the methodology, provided supervision and wrote the final draft.

\section{Acknowledgement}

We thank Emmanuel Ameisen and Johnny Lin  for their helpful discussion and feedback on the initial experiments. We are thankful to Open Philanthropy for funding the work of Gonçalo Paulo. We are grateful to CoreWeave for providing part of the compute resources.

\bibliographystyle{icml2026}
\bibliography{refs}

\appendix
\onecolumn
\section{Reproducibility Statement}
\label{sec:repro}

To support the workshop's emphasis on reproducibility and code/data access, we release the following anonymously for the review period.

\begin{itemize}\itemsep0pt
  \item \textbf{Code.} Repository at \url{https://github.com/peppinob-ol/attribution-graph-probing} (commit hash and license in \texttt{README.md}).
  \item \textbf{Datasets.} The full 44{,}596-run swap dataset (graphs, prompts, activations, supernode groupings, swap dumps for all five domains) at the same  URL, with a per-domain manifest in \cref{appx:repro}.
  \item \textbf{Interactive demo.} An anonymized HuggingFace Space at \url{https://huggingface.co/spaces/Peppinob/concept-swap-explorer} that showcases concept swaps across 82 entities in 4 domains.
  \item \textbf{Per-domain CLI examples.} \cref{appx:repro} gives end-to-end commands for the USA, Books, Products, Paintings, Sounds pipelines, including the exact $\tau$, $M$, and seed used for each table and figure.
  \item \textbf{Determinism.} The pipeline is fully deterministic (sha256-seeded random controls; fixed temperature/seed/penalty for generation; no random initialization); per-feature checkpoints permit resumption from partial runs.
  \item \textbf{Hardware.} A single L4 GPU is sufficient to reproduce CPAS for any single entity in 14--24 minutes (graph generation $+$ activation measurement $+$ grouping $+$ subgraph), excluding the swap sweep; the full swap sweep is documented per-domain in \cref{appx:repro}.
\end{itemize}


\section*{Appendix}
\addcontentsline{toc}{section}{Appendix: How to Read}

The appendix is organized to track the main paper: \S\ref{appx:pipeline}--\S\ref{appx:emblematic} give a detailed reference for the pipeline, its functional vocabulary, and emblematic feature examples per role; \S\ref{appx:case} collects worked steering case studies; \S\ref{appx:fieldadd}--\S\ref{appx:kl} back up the field-additivity, $M$-search, and steering-strength discussions of \S\ref{sec:results}; \S\ref{appx:regime} gives the logit-shift regime taxonomy used throughout; \S\ref{appx:clustering} validates the four-role partition against the natural cluster geometry of the CPAS metric space; \S\ref{appx:dallas_case} is the human-curated Dallas case study with comparison with our pipeline and a fixed-$K$ saturation control; \S\ref{appx:specificity}--\S\ref{appx:topkim} give per-domain labeled-vs-random and per-pair influence-matched top-$K$ results; \S\ref{appx:suppression} details the observation that suppression-is-easy, steering-is-hard \S\ref{appx:crossprompt}--\S\ref{appx:scaffold} back up the cross-prompt and scaffold analyses; \S\ref{appx:negative} discusses the weakest of our four domains (Paintings); \S\ref{appx:repro} is the reproducibility manifest; \S\ref{appx:glossary} is the glossary. Every appendix section is self-contained and cross-referenced from the main text.

\section{Pipeline Reference}
\label{appx:pipeline}

\subsection{End-to-end pipeline}

The pipeline takes an attribution graph for a seed prompt and returns a collection of concept-aligned supernodes together with the data needed to run matched-control swap interventions. The pseudocode below summarizes the steps; the paragraphs that follow describe each step in plain language.

\begin{small}
\begin{verbatim}
Input:  attribution graph G_p, seed prompt p, target logit y
Output: concept-aligned supernodes {S_i}, swap harness I

1. V_tau  = select_features_by_cumulative_influence(G_p, tau=0.95)
2. C      = llm_generate_concepts(p, K)
3. Probes = synthesize_probes(C, template_constraints)
4. for each feature f in V_tau, each probe q in Probes:
     A[f, q] = measure_activation(f, q)
5. CPAS[f] = aggregate_cross_prompt(A[f, *])
6. for each feature f in V_tau:
     role[f] = classify(CPAS[f], thresholds)
     name[f] = assign_name(f, A, blacklist, target_token_rules)
7. Supernodes = group_by((role, name)), stability >= 0.6
8. for each swap pair (e_A, e_B):
     I_lab  = build_swap(S(e_A), S(e_B), M_ablate=-2, M_amplify=20)
     I_rand = matched_random_control(I_lab)
     run(I_lab); run(I_rand)
9. adaptive_M_search(missed_pairs)
\end{verbatim}
\end{small}

\emph{Step 1 (feature selection).} Attribution graphs from Neuronpedia are generated with a relatively permissive node-inclusion threshold (\texttt{nodeThreshold=0.8}), a high edge-influence threshold (\texttt{edgeThreshold=0.85}), and a cap of \texttt{maxFeatureNodes=5000}. The graph typically contains 600--5{,}000 nodes; we then retain only the nodes whose cumulative influence sums to at least $\tau{=}0.95$, which reduces each circuit to 200--700 features for batch experiments.

\emph{Step 2--3 (probe generation).} The seed prompt is handed to a separate LLM together with a small specification that (a) lists the \emph{concepts} we expect to find in the graph (entities, relations, answer tokens) and (b) constrains the probe templates to be syntactically close to the seed (same prepositions, same answer position, same length). The LLM returns a handful of concept-targeted probe prompts. Typical probe counts are 4--5 per seed.

\emph{Step 4--5 (activation measurement and CPAS).} Each feature in $V_\tau$ is re-run on each probe and its activation measured with the Neuronpedia API. The per-(feature, probe) record stores the peak token, the peak position in the probe, the activation density, a sparsity ratio, a robust $z$-score, and a cosine similarity to the seed activation. These raw measures are then aggregated across probes into a compact per-feature signature (the CPAS).

\emph{Step 6 (classification and naming).} A strict-priority decision tree (\S\ref{sec:method:rules}; thresholds in \Cref{tab:decision-thresholds}) assigns each feature one of four functional roles (Semantic-Dictionary, Semantic-Concept, Relationship, Say-X) or a fifth \textsc{Review} bucket that is excluded from downstream analysis. Naming is role-specific (\S\ref{sec:method:supernodes}).

\emph{Step 7 (supernode formation).} Same-role same-name features are merged into a supernode; features whose role or name is inconsistent across probes ($<60\%$ probe-level stability) are marked \emph{ungrouped} and excluded.

\emph{Step 8--9 (causal-validation harness).} For every entity pair in the domain we build an additive entity-swap intervention (\S\ref{sec:method:harness}), run it at $M_{\mathrm{ablate}}{=}-2$ / $M_{\mathrm{amplify}}{=}20$ with attention \emph{not} frozen, and compare it to three deterministic matched-random-control replicates (\S\ref{sec:method:harness}). Pairs that miss at the defaults enter adaptive $M$-search (\S\ref{appx:msearch}).

\subsection{CPAS measures}
The CPAS is the seven-number summary that the decision tree operates on. \Cref{tab:cpas-metrics} gives the formal definitions and the intuition for each field.

\begin{table}[h]
\centering\small
\begin{tabular}{p{0.28\linewidth}p{0.64\linewidth}}
\toprule
Field & Definition and intuition \\
\midrule
\texttt{peak\_consistency\_main} & Fraction of probes whose peak token matches the feature's modal peak. High ($\geq 0.8$) means the feature is a dictionary-like detector for one particular token. \\
\texttt{n\_distinct\_peaks} & Number of distinct peak tokens across \emph{active} probes. $=1$ for a Semantic-Dictionary feature; larger for Semantic-Concept or Say-X. \\
\texttt{share\_F} / \texttt{conf\_F} & Share of active probes whose peak falls on a functional token / classifier confidence on that share. A feature with \texttt{share\_F}$\approx 1$ and high \texttt{conf\_F} is a Say-X candidate. \\
\texttt{func\_vs\_sem\_pct} & $100 \cdot \big(\max_{\text{func}} - \max_{\text{sem}}\big) / \max_{\text{overall}}$. Positive when the peak is on a functional token; used to break ties between Say-X and semantic roles. \\
\texttt{sparsity\_median} & Median of $(\mathrm{peak} - \mathrm{mean})/\mathrm{peak}$ across active probes. Low values indicate a dense, diffuse activation pattern characteristic of Relationship features. \\
\texttt{conf\_S} & $1 - \mathrm{share\_F}$, i.e.\ confidence that the peak is on a semantic token. \\
\bottomrule
\end{tabular}
\caption{CPAS fields used by the classification rules in \S\ref{sec:method:rules}.}
\label{tab:cpas-metrics}
\end{table}

\subsection{Decision-rule thresholds}

The decision tree is a strict-priority cascade: features are evaluated against the rules in the order below, and the first matching rule wins. Thresholds were iteratively refined on a small set of pilot circuits and then frozen before the 44{,}596-run sweep.

\begin{table}[h]
\centering\small
\begin{tabular}{lll}
\toprule
Priority & Role & Rule \\
\midrule
1 & \textsc{Sem-Dict} & \texttt{peak\_consistency\_main} $\geq 0.80$ \textsc{and} \texttt{n\_distinct\_peaks} $\leq 1$ \\
2 & \textsc{Say-X}    & \texttt{func\_vs\_sem\_pct} $\geq 50$ \textsc{and} \texttt{conf\_F} $\geq 0.90$ \textsc{and} layer $\geq 7$ \\
3 & \textsc{Rel}      & \texttt{sparsity\_median} $< 0.45$ \\
4 & \textsc{Sem-Conc} & layer $\leq 3$ \textsc{or} \texttt{conf\_S} $\geq 0.50$ \textsc{or} \texttt{func\_vs\_sem\_pct} $< 50$ \\
5 & \textsc{Review}   & anything not matching the above (excluded) \\
\bottomrule
\end{tabular}
\caption{Classification thresholds used for the four functional roles (\S\ref{sec:method:rules}). Strict priority: the first matching rule wins.}
\label{tab:decision-thresholds}
\end{table}

The layer-based asymmetry (Say-X requires layer $\geq 7$, Sem-Conc accepts layer $\leq 3$) encodes a rough prior observed in Gemma-2-2B: early layers are overwhelmingly semantic-content features; the "Say X" behavior of functional-token features promoting a specific answer is observed only in the later half of the network.

\paragraph{Dictionary subtype: strict and fallback.}
The semantic role is split internally into three \emph{subtypes} that determine downstream naming. Rule~1 (\texttt{peak\_consistency\_main} $\geq 0.80$ and \texttt{n\_distinct\_peaks} $\leq 1$) produces the strict \emph{Dictionary} subtype: stable single-token detectors. Rule~4 produces either \emph{Concept} (the \texttt{conf\_S} or \texttt{func\_vs\_sem\_pct} branch) or, when fired through its low-layer branch (\texttt{layer} $\leq 3$), the \emph{Dictionary (fallback)} subtype: early-layer semantic features whose peak token is not stable enough to satisfy Rule~1 but whose layer prior pulls them into a dictionary-like reading as the safest default. The two dictionary subtypes are reported together as \textsc{Sem-Dict} in every per-domain table; only the cluster-geometry analysis (\S\ref{appx:clustering}) keeps them separate, which is where the \emph{Dictionary (fallback)} bucket reveals additional internal structure.

\section{Functional Vocabulary and Target-Token Mapping}
\label{appx:funcvocab}

The pipeline distinguishes \emph{functional} tokens (copulas, articles, prepositions, conjunctions, relative pronouns, auxiliaries) from \emph{semantic} tokens (content-bearing), and uses this distinction both in the Say-X rule and during supernode naming.

\paragraph{English functional vocabulary.}
We use a fixed list of 87 English functional tokens. The groups and their members are:

\begin{itemize}\itemsep0pt
\item \textbf{Copulas and copular contractions}: \texttt{is, was, are, were, be, been, being, am, 's, 're, 'm}.
\item \textbf{Articles and demonstratives}: \texttt{the, a, an, this, that, these, those}.
\item \textbf{Prepositions}: \texttt{of, in, on, at, to, for, with, by, from, as, into, onto, upon, about, above, below, between, among, through, during, before, after}.
\item \textbf{Conjunctions}: \texttt{and, or, but, nor, so, yet, for}.
\item \textbf{Relative pronouns and wh-words}: \texttt{that, which, who, whom, whose, where, when, why, how}.
\item \textbf{Auxiliaries}: \texttt{do, does, did, have, has, had, will, would, shall, should, can, could, may, might, must}.
\end{itemize}

\paragraph{Directionality of the $\pm 7$-token search.}
When a feature peaks on a functional token, we search the probe for the nearest \emph{semantic} target in a window of $\pm 7$ tokens. The direction of the search depends on the functional token's syntactic role: copulas, auxiliaries, articles, and most prepositions look \emph{forward} (the semantic target is typically the noun phrase they introduce); \texttt{of} and possessive \texttt{'s} look \emph{backward} (the semantic target is the head noun that governs the functional token); conjunctions and punctuation are searched \emph{bidirectionally}, nearest-first. Multi-token semantic targets resolve to the first subword.

\paragraph{Cross-lingual behavior and a documented failure mode.}
We observed one clean cross-lingual failure during development, on a French version of the antonym seed ``le contraire de `petit' est''. Running the English pipeline on a French prompt produced incoherent cross-prompt activations: functional tokens were often peaked on the English contraction \texttt{'s} or on French tokens (\texttt{de}, \texttt{est}) that were not in our vocabulary, so the Say-X rule either never fired or fired on the wrong side. Adding a small French functional vocabulary and expanding the blacklist (bridging tokens that appear in both languages) restored reliable grouping. We flag this as a concrete instance of the cross-architecture-and-language limitation discussed in \S\ref{sec:limitations}.

\section{Feature examples per Functional Role}
\label{appx:emblematic}

To make the four functional roles of \S\ref{appx:pipeline} concrete, we hand-pick one canonical feature per role---five panels, since \textsc{Sem-Dict} is shown both in its strict and fallback subtype. Each panel stacks the seed prompt (top, bordered) with the five probe prompts of the entity's CPAS dump; cell shading is the per-token activation normalized by the feature's global maximum across all probes, numbers are the raw activations, and values below the visibility threshold are left blank. Candidates were selected from the cross-dataset feature manifest and inspected manually; the metrics quoted in each caption are the raw values that placed the feature inside its branch of the decision tree of \cref{tab:decision-thresholds}.

\begin{figure}[h]
  \centering
  \includegraphics[width=0.95\linewidth]{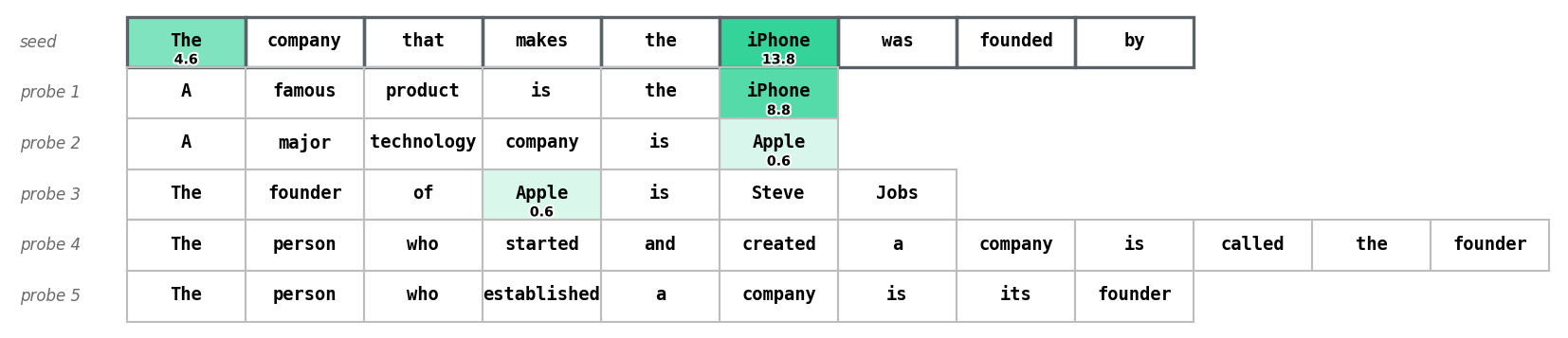}
  \caption{\textsc{Sem-Dict} (strict) --- \emph{iPhone} entity, products domain, layer-0 feature 1285. The feature lights up on the literal token \texttt{iphone} in every probe and is silent almost everywhere else: the same surface token, every time, with no spread to morphological variants or related concepts. This is a canonical single-token detector that sits at the strict end of the dictionary branch of the decision tree.}
  \label{fig:emblematic-semdict-strict}
\end{figure}

\begin{figure}[h]
  \centering
  \includegraphics[width=0.95\linewidth]{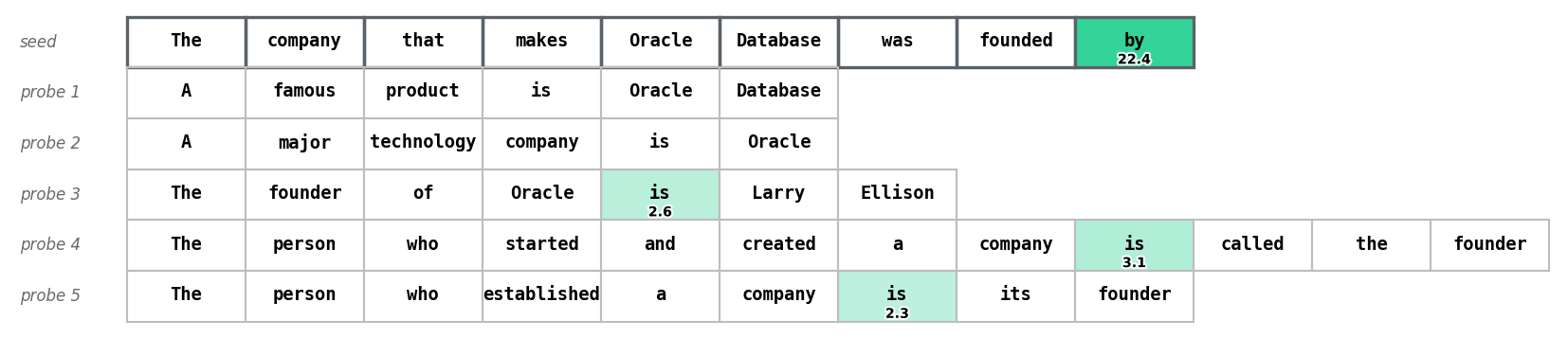}
  \caption{\textsc{Say-X} --- \emph{Say (founder)} entity, products domain, layer-11 feature 9574. A late-layer feature whose only strongly-active position in each probe is the copula \texttt{is} immediately preceding \texttt{selected} next-token prediction. Because every active position is a functional token, the $\pm 7$-token search of \S\ref{appx:funcvocab} resolves the actual semantic target by looking forward from the copula---in this case, into the answer slot.}
  \label{fig:emblematic-sayx}
\end{figure}

\begin{figure}[h]
  \centering
  \includegraphics[width=0.95\linewidth]{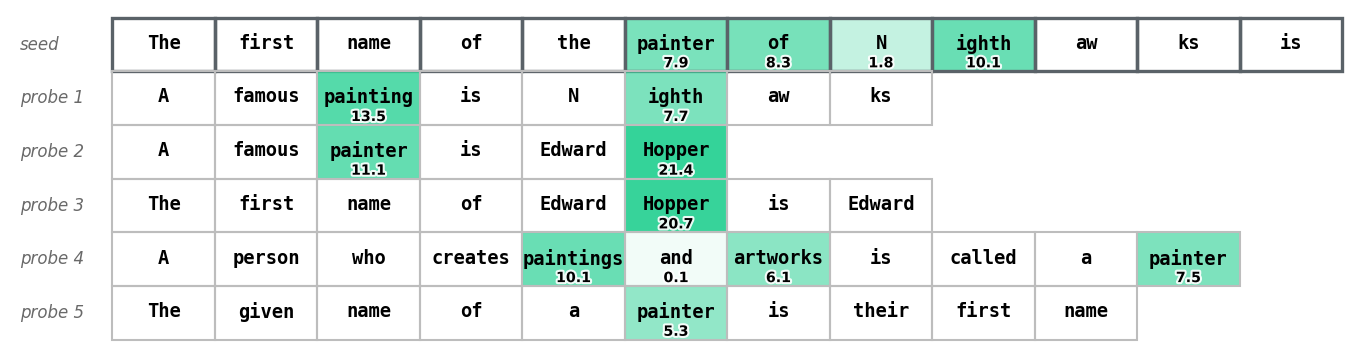}
  \caption{\textsc{Sem-Conc} --- \emph{Hopper} entity, paintings domain, layer-4 feature 47039. The same feature fires on \emph{Edward}, \emph{Hopper} and on the painting tokens (\emph{Nighthawks}) across all five probes, distributing its activation across several distinct semantic peaks anchored on a single coherent concept (the artist Edward Hopper). Multi-peak coverage of a concept is the visual signature that distinguishes \textsc{Sem-Conc} from the strict-dictionary case above.}
  \label{fig:emblematic-semconc}
\end{figure}

\begin{figure}[h]
  \centering
  \includegraphics[width=0.95\linewidth]{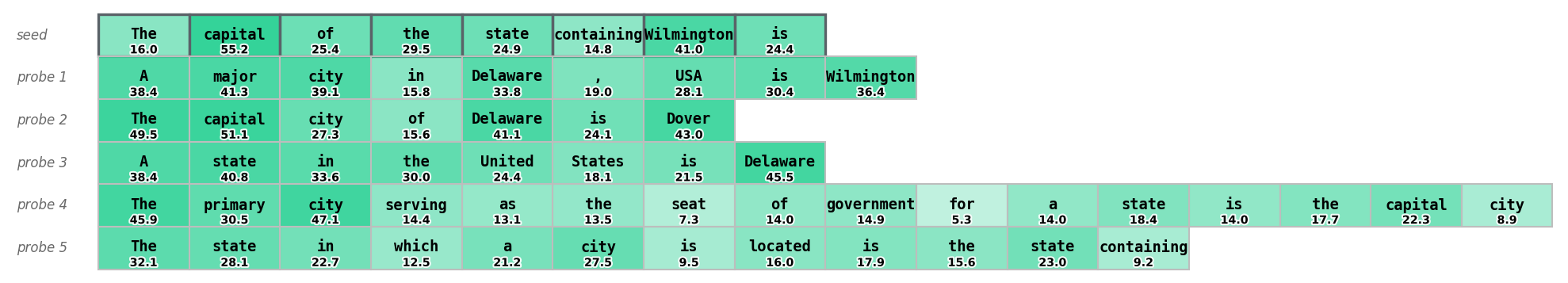}
  \caption{\textsc{Relationship} --- \emph{(Capital) related} entity, U.S.\ states domain, layer-7 feature 66851. The feature is active on a substantial fraction of every probe: a diffuse green band runs across each row, with several peaks spread between functional and semantic positions and no single token dominating the activation budget. This combination---low sparsity, no clear peak target---is the visual signature of \textsc{Relationship} features. Major peaks (\emph{capital}, \emph{Delaware}, \emph{Wilmington}, \emph{state}) inform possible main links.}
  \label{fig:emblematic-relationship}
\end{figure}

\begin{figure}[h]
  \centering
  \includegraphics[width=0.95\linewidth]{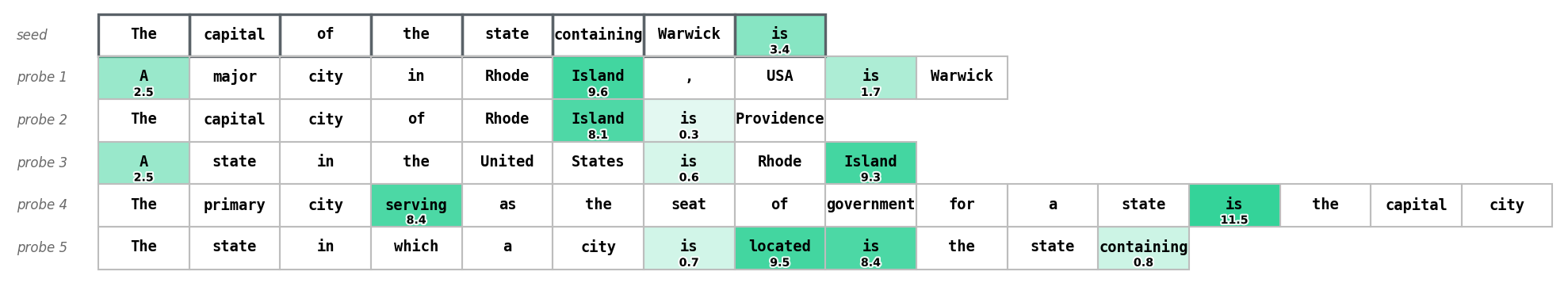}
  \caption{\textsc{Sem-Dict} (fallback) --- \emph{Island} entity, U.S.\ states domain, layer-3 feature 81381. The feature fires on every probe and behaves like a Rhode-Island detector, but its activation does not collapse onto a single surface token: it lights up both on a subword of the multi-token state name (\texttt{island}) and re-fires on functional tokens and general concepts. The fallback rule of \S\ref{appx:pipeline} is meant to capture  this regime---dictionary-like specificity at a layer too shallow for the semantic concept criterion, with a peak that drifts across morphologically related tokens before the network has had room to collapse them onto a single representative.}
  \label{fig:emblematic-semdict-fallback}
\end{figure}


\section{Steering Case Studies}
\label{appx:case}

This section collects worked case studies that would be too discursive for the main text. Each entry gives the source and target entity, the field-additivity variant used, the intervention size, the vsMax achieved, and the first $\sim$10 generated tokens of the model's continuation.

\subsection{Five success cases}

\begin{figure*}[!htbp]
  \centering
  \includegraphics[width=\linewidth]{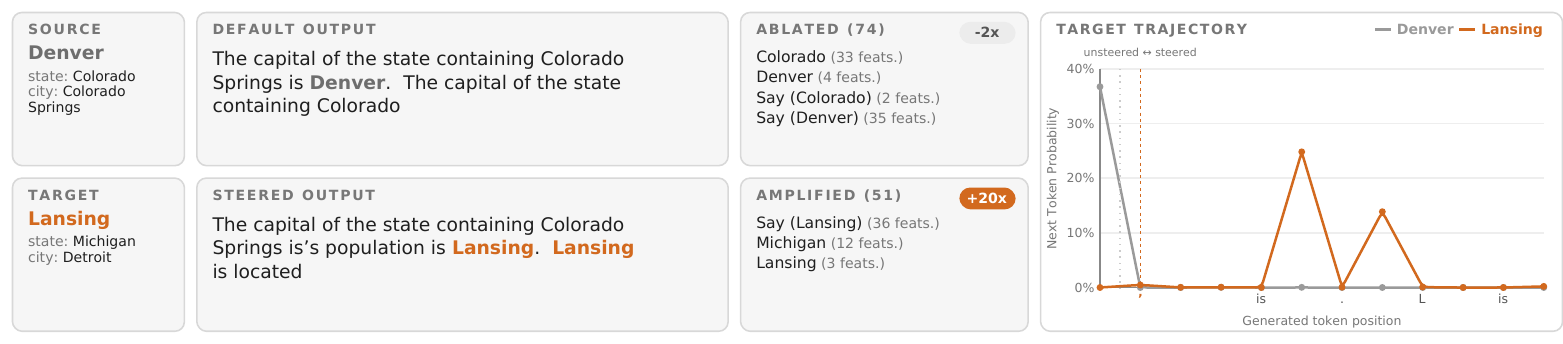}
  \caption{Colorado Springs $\to$ Detroit (\texttt{state+capital} variant). Steered ``Lansing'' wins at trajectory positions 4 and 6 (peak $\sim$25\%) while the default ``Denver'' attractor is fully suppressed; the supporting Michigan and Say(Lansing) supernodes lift the answer without disturbing the prompt-anchor token. vsMax $+17.5$, target logit rank $342 \to 1$. Regime~A in logit trajectory (target UP, source DOWN), clean flip, on-topic continuation.}
  \label{fig:case-colorado-detroit}
\end{figure*}

\begin{figure*}[!htbp]
  \centering
  \includegraphics[width=\linewidth]{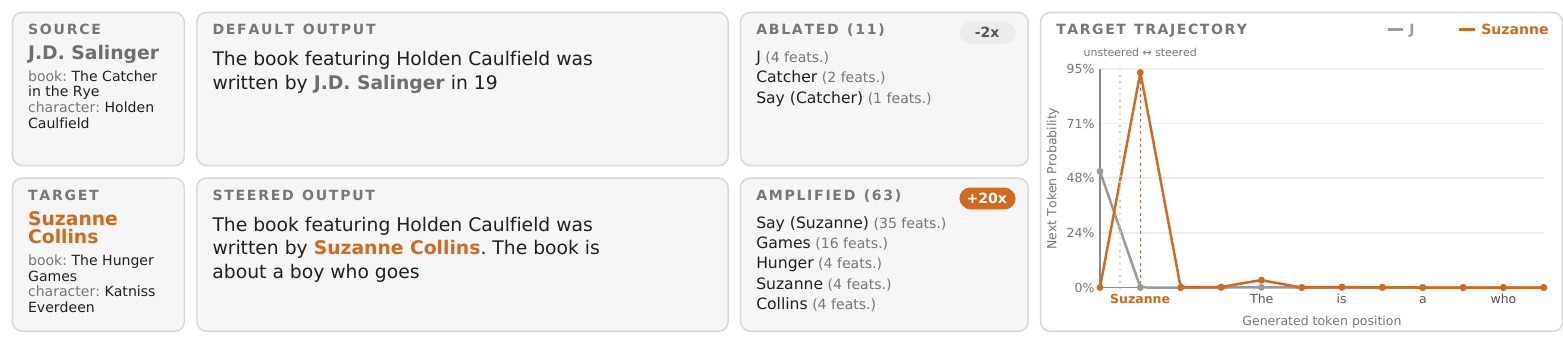}
  \caption{Holden Caulfield $\to$ Katniss Everdeen (\texttt{book+author} variant). Eleven ablated features remove ``J.D.''; 63 amplified features push ``Suzanne'' to $p \approx 0.95$ at position~0, and the steered continuation also recovers the correct book title (``The Hunger Games''). The amplify panel is dominated by Say(Suzanne) and a Hunger-Games scaffold of $\sim 16$+$4$ features. vsMax $+21.79$, the largest single-pair vsMax in our dataset; illustrates how a small but correctly-targeted amplify set can dominate a graph $\sim 50\times$ its size.}
  \label{fig:case-holden-katniss}
\end{figure*}

\begin{figure*}[!htbp]
  \centering
  \includegraphics[width=\linewidth]{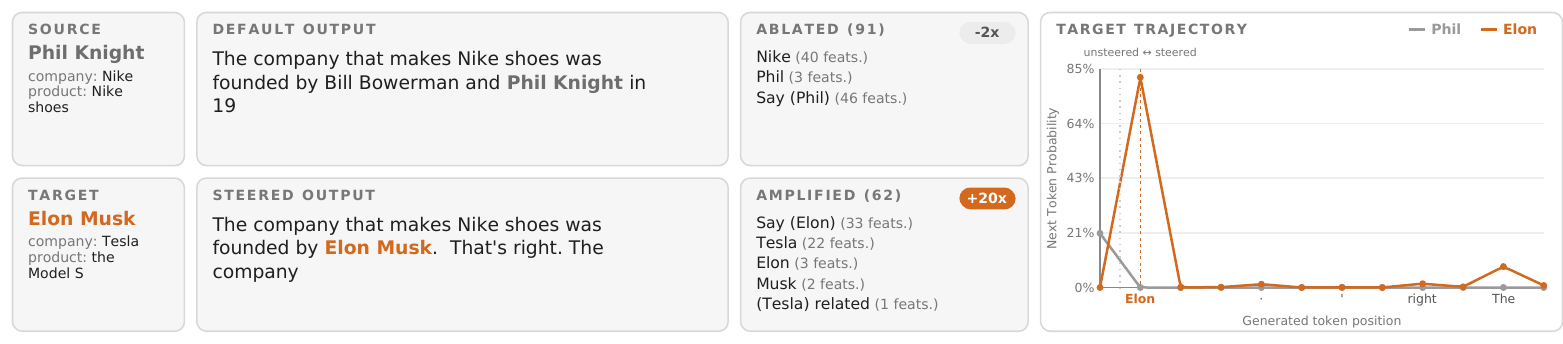}
  \caption{Nike $\to$ Tesla (\texttt{company+founder} variant). The orange ``Elon'' curve spikes to $\sim 85\%$ at position~0 and the grey ``Phil'' source vanishes immediately; the steered continuation also produces the correct company token ``Tesla''. The 91 ablated features are dominated by a single Nike supernode (40 features). vsMax $+12.25$, target rank $5{,}933 \to 1$, steered first-token probability $p{=}0.82$. Redirection is complete at position~0.}
  \label{fig:case-nike-tesla}
\end{figure*}

\begin{figure*}[!htbp]
  \centering
  \includegraphics[width=\linewidth]{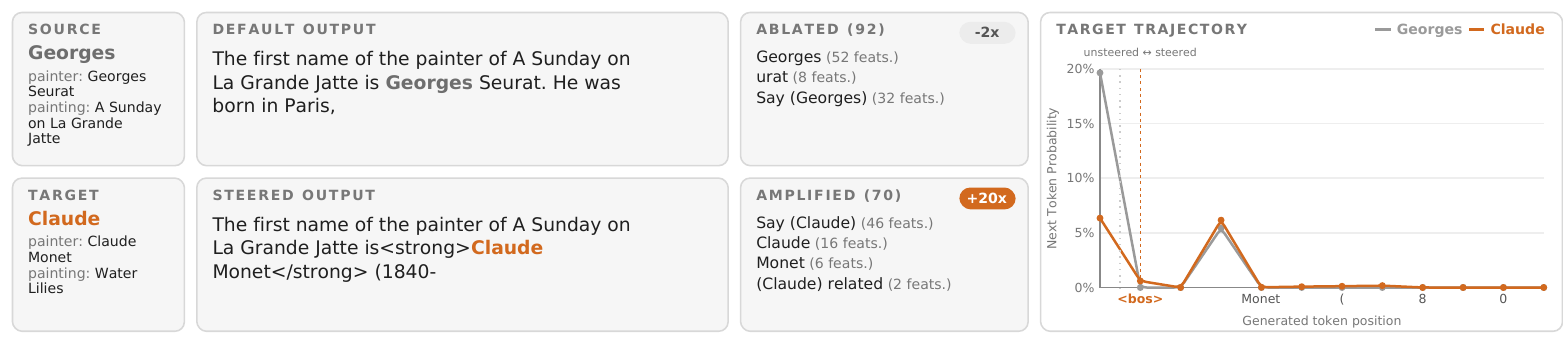}
  \caption{La Grande Jatte $\to$ Water Lilies (\texttt{painter+first\_name} variant). The orange ``Claude'' curve barely peaks at $\sim 6\%$; ``Georges'' still dominates the trajectory and the steered continuation actually could re-emit ``Georges Seurat''. A Hit in name only -- the position-0 logit prefers Claude. vsMax $+9.375$; The baseline rank of ``Claude'' was already~$3$. Illustrates that Paintings can be redirected only when baseline conditions are already favourable.}
  \label{fig:case-jatte-lilies}
\end{figure*}

\begin{figure*}[!htbp]
  \centering
  \includegraphics[width=\linewidth]{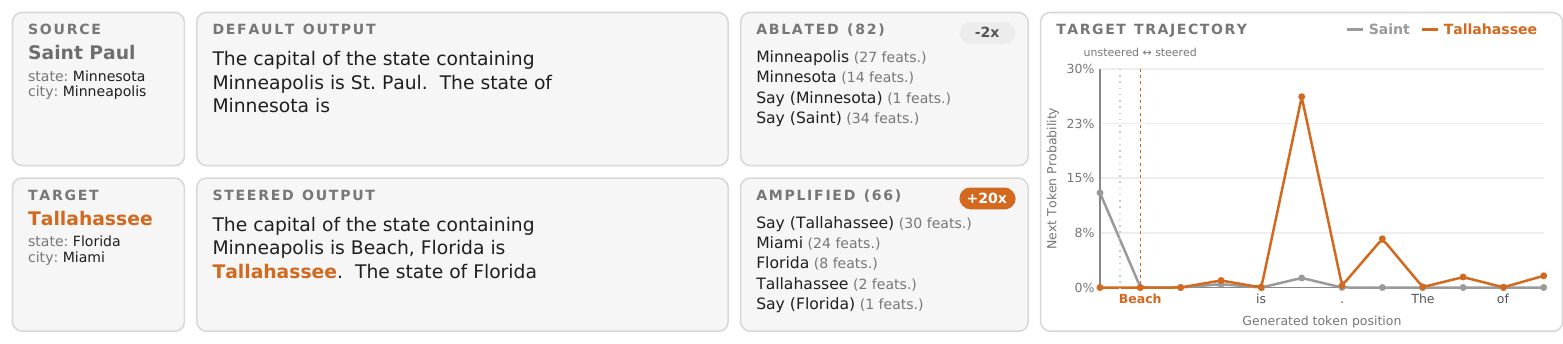}
  \caption{Minnesota $\to$ Florida (all-three-fields variant, \texttt{state+capital+city}, $M{=}20$). Position~0 emits the incoherent token ``Beach''; ``Tallahassee'' only spikes to $\sim 30\%$ at position~3, after the autoregressive decoder has already committed to a derailed continuation. The grey source curve never disappears, signalling residual leakage from the un-ablated source scaffold. vsMax $+16.06$ but flip@0 is \texttt{False}, and control-stability reaches $16.89$ (very high collateral disruption). Clear illustration of the all-fields cost: position-0 disruption corrupts the first generated token even though the target logit eventually wins later.}
  \label{fig:case-minnesota-florida}
\end{figure*}

\FloatBarrier

\subsection{Six failure cases covering four of the five failure modes}

\begin{figure*}[!htbp]
  \centering
  \includegraphics[width=\linewidth]{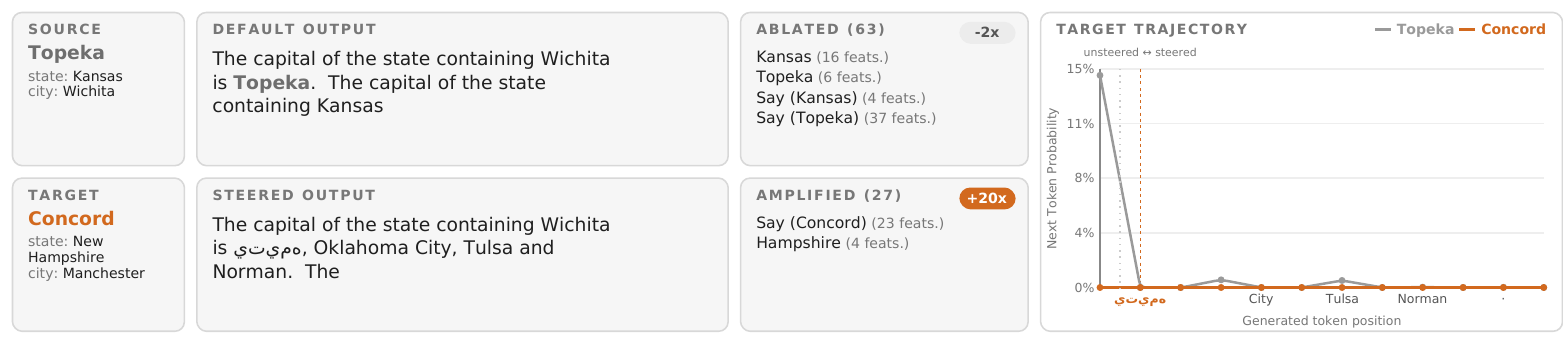}
  \caption{Kansas/Wichita $\to$ New Hampshire/Manchester (total failure). The amplified target features fail to lift ``Concord'' over the dominant ``Oklahoma'' attractor; the steered continuation collapses into a different neighbouring-state capital (``Oklahoma City\ldots Tulsa''). Trajectory shows neither curve ever crosses $\sim 10\%$ -- a complete failure to retarget. vsMax $-6.875$, target rank $5{,}816$. Kansas source features are successfully suppressed, but the amplified New Hampshire features do not encode Manchester-capital circuitry strongly enough. Failure mode: feature specificity.}
  \label{fig:case-kansas-nh}
\end{figure*}

\begin{figure*}[!htbp]
  \centering
  \includegraphics[width=\linewidth]{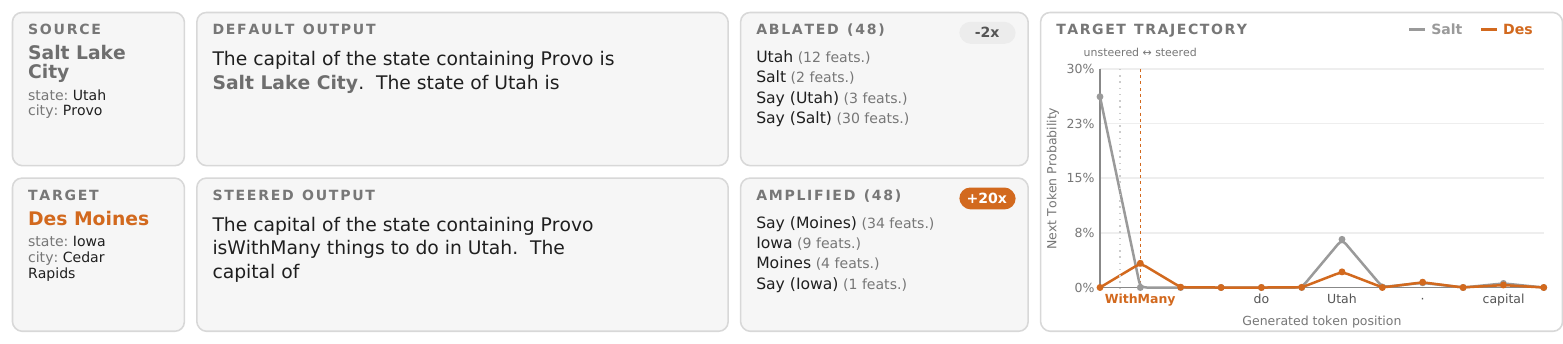}
  \caption{Utah/Provo $\to$ Iowa/Cedar Rapids (rank-hit misalignment). Textbook RkGrp-without-Hit case: the trajectory shows the target winning at position~0, but the steered output reverts to a Utah-themed ramble rather than naming Des Moines anywhere. vsMax $+14.0$, RkGrp $1$; the target logit wins at position~0 but subsequent tokens are disrupted enough that the Hit criterion (target token anywhere in continuation) fails. Failure mode: feature-interaction noise.}
  \label{fig:case-utah-iowa}
\end{figure*}

\begin{figure*}[!htbp]
  \centering
  \includegraphics[width=\linewidth]{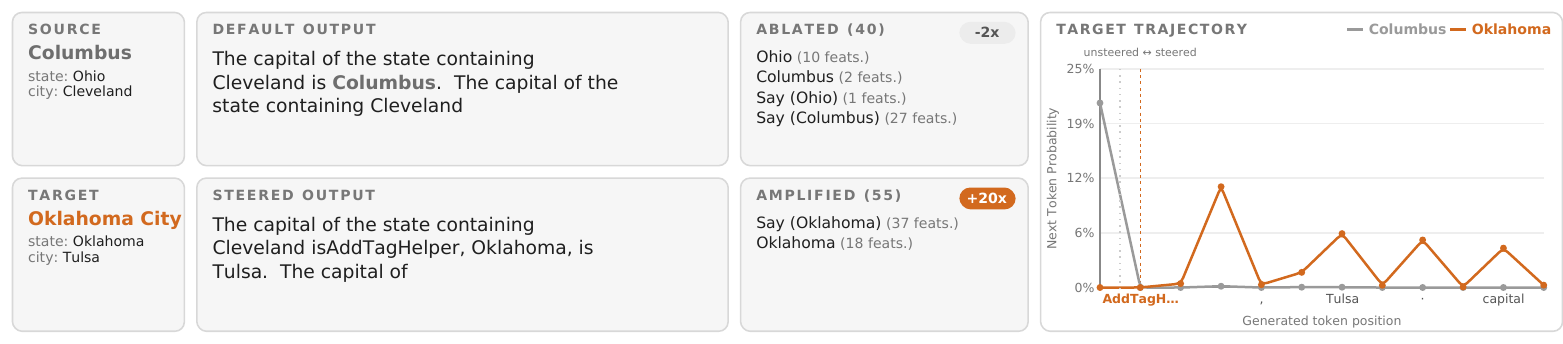}
  \caption{Ohio/Cleveland $\to$ Oklahoma/Tulsa (regime~D). Position~0 emits ``AddTagHelper'' (a corrupted code-token), poisoning the autoregressive decoder; the orange ``Tulsa'' curve does rise later but never overtakes the grey source, so by the Hit criterion the intervention fails despite a positive vsMax. vsMax $+13.31$ but flip@0 is \texttt{False}; at $M{=}5$ this same pair becomes a Hit. Failure mode: severe overshoot, rescuable by $M$-reduction.}
  \label{fig:case-ohio-oklahoma}
\end{figure*}

\begin{figure*}[!htbp]
  \centering
  \includegraphics[width=\linewidth]{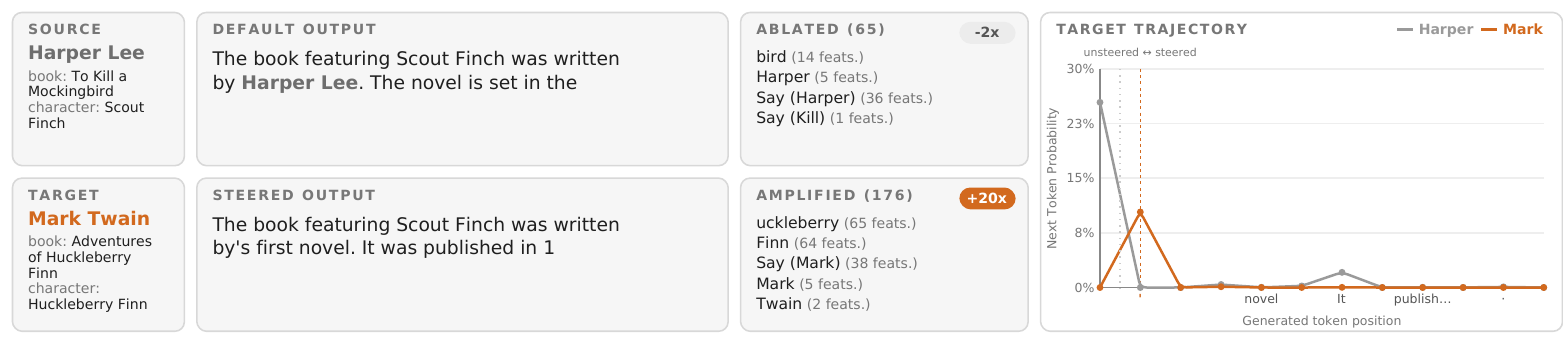}
  \caption{Scout Finch $\to$ Huckleberry Finn (\texttt{book+author} variant; field interference / substring confound). The amplified ``Finn'' supernode (64 features) and ``uckleberry'' (65 features) interfere with each other; the steered output produces neither ``Mark Twain'' nor a coherent book title, but a possessive fragment (``'s first novel''). The orange ``Mark'' peak at $\sim 8\%$ is dwarfed by the noise on adjacent positions. 176 amplify features (the highest count in Books), with a substring confound (\texttt{Finn} is a common suffix). Rescuable at the \texttt{book+author} variant but not at the all-three-fields baseline. Failure mode: field interference.}
  \label{fig:case-scout-huck}
\end{figure*}

\begin{figure*}[!htbp]
  \centering
  \includegraphics[width=\linewidth]{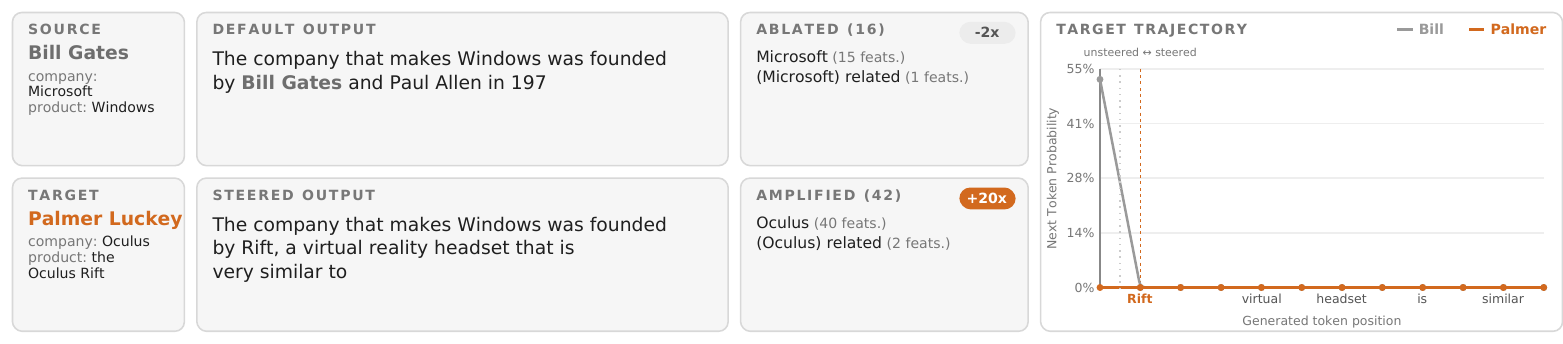}
  \caption{Windows $\to$ Oculus (\texttt{company}-only variant; product-identity collapse). With only 16 ablate / 42 amplify features, the trajectory shows ``Rift'' at $\sim 55\%$ at position~0; the orange ``Palmer'' curve never lifts off zero. The steered continuation reads as a product description rather than a founder attribution. The source (Windows/Microsoft) is suppressed cleanly but the ``company'' supernodes in the Products domain encode product identity; amplifying them therefore promotes the product name. Failure mode: feature specificity---a perfectly-executed intervention that targets the wrong entity because the label's extension does not match the target concept.}
  \label{fig:case-windows-oculus}
\end{figure*}

\begin{figure*}[!htbp]
  \centering
  \includegraphics[width=\linewidth]{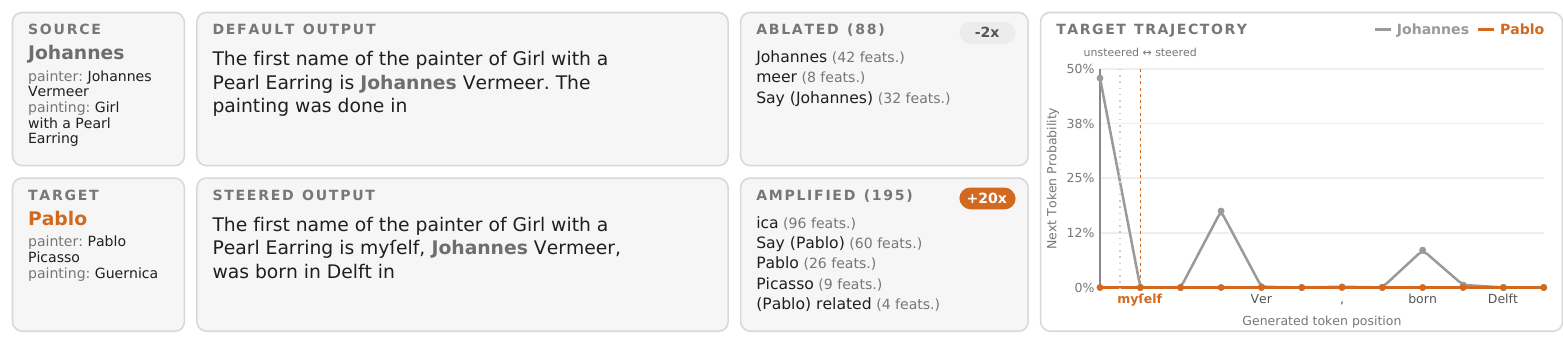}
  \caption{Girl with a Pearl Earring $\to$ Guernica (\texttt{painter+first\_name} variant; cross-concept collapse). Even at 195 amplified features the orange ``Pablo'' curve never crosses $\sim 5\%$; the grey ``Johannes'' source still spikes to $\sim 50\%$ at position~2 and to $\sim 13\%$ at position~5, confirming the Vermeer circuit re-emerges in the steered continuation despite source ablation. 195 amplify and 88 ablate features fail to dislodge the source painter; ``Pablo'' ends up at rank $1{,}382$ at position~0. The source circuit is deeply entrenched (Vermeer features are shared across the painting and its description) and the target is coarse (Picasso has many broader associations). Failure mode: feature specificity combined with low scaffold compatibility.}
  \label{fig:case-girl-guernica}
\end{figure*}

\FloatBarrier

\subsection{Threats to validity for case studies}
Case studies select for extremes (the largest vsMax pairs and the most striking failures), and single-example interpretations should not be over-generalized. The distribution-level claims live in the full labeled-vs-random tables (\S\ref{appx:specificity}), field-additivity tables (\S\ref{appx:fieldadd}), and adaptive-$M$ rescue rates (\S\ref{appx:msearch}). Post-hoc explanations of success and failure (e.g.\ ``the company supernodes encode product identity'') are hypotheses supported by the particular pair.



\section{Field-Additivity Detail}
\label{appx:fieldadd}

Each domain has three semantic fields: an \emph{input} field (the entity being named in the prompt), an \emph{intermediate} field (the bridging concept, e.g.\ the US state), and an \emph{answer} field (the concept whose logit we measure). For each entity pair the field-additivity sweep runs seven intervention variants: three singletons, three pairs, and the all-three baseline. The bolded row is the best variant by Hit\%.

\paragraph{USA (state, capital, city; 2{,}450 pairs/variant).}
\begin{center}\small
\begin{tabular}{lrrrr}
\toprule
Subset & Hit\% & vsMax & RkGrp & MedRk \\
\midrule
city                &  8.3 & $-1.26$ & 6.92 & 190 \\
state               & 24.1 & $+2.38$ & 2.72 & 14 \\
capital             & 19.8 & $+0.63$ & 3.69 & 11 \\
\textbf{state+capital} & \textbf{43.6} & \textbf{+4.00} & \textbf{1.47} & \textbf{3} \\
state+city          & 15.8 & $+1.71$ & 2.31 & 28 \\
capital+city        & 15.9 & $+0.58$ & 4.16 & 30 \\
all 3               & 31.5 & $+2.86$ & 1.72 & 5 \\
\bottomrule
\end{tabular}
\end{center}

Prompts are of the form ``The capital of the state containing $\langle$city$\rangle$ is''. The input field is the city (what the model reads), the intermediate is the state (the reasoning bridge), and the answer is the capital. The state+capital pair gives the best redirection; adding the city field degrades every metric---the mechanism we label \emph{less is more}: city features encode what the model reads.

\paragraph{Books (character, book, author; 90 pairs/variant).}
\begin{center}\small
\begin{tabular}{lrrrr}
\toprule
Subset & Hit\% & vsMax & RkGrp & MedRk \\
\midrule
character        &  8.9 & $+1.87$ & 1.77 & 125 \\
book             & 63.3 & $+6.69$ & 1.20 & 2 \\
author           & 10.0 & $+4.71$ & 1.33 & 12 \\
\textbf{book+author} & \textbf{66.7} & \textbf{+7.76} & \textbf{1.02} & \textbf{2} \\
character+book   & 15.6 & $+5.22$ & 1.23 & 24 \\
character+author &  4.4 & $+3.45$ & 1.30 & 76 \\
all 3            &  5.6 & $+5.97$ & 1.03 & 17 \\
\bottomrule
\end{tabular}
\end{center}

Prompts are of the form ``The author of the book featuring $\langle$character$\rangle$ is''. The character field is the input; intervening on it alone moves almost nothing. The largest single-field effect comes from the book field ($63.3\%$ Hit on its own), and adding the author field only nudges the total up further ($66.7\%$). Adding the character field undoes the effect almost entirely ($5.6\%$ Hit for all-three). This is the sharpest single-domain illustration of the less-is-more effect.

\paragraph{Products (product, company, founder; 132 pairs/variant).}
\begin{center}\small
\begin{tabular}{lrrrr}
\toprule
Subset & Hit\% & vsMax & RkGrp & MedRk \\
\midrule
product              &  0.8 & $+1.90$ & 1.61 & 137 \\
company              &  9.1 & $+2.62$ & 1.39 & 128 \\
founder              & 16.7 & $+2.08$ & 1.27 & 18 \\
\textbf{company+founder} & \textbf{34.1} & \textbf{+3.06} & \textbf{1.27} & \textbf{18} \\
product+company      &  2.3 & $+2.78$ & 1.31 & 93 \\
product+founder      &  3.8 & $+2.54$ & 1.23 & 48 \\
all 3                & 23.5 & $+3.47$ & 1.20 & 26 \\
\bottomrule
\end{tabular}
\end{center}

\paragraph{Paintings (painting, painter, first\_name; 90 pairs/variant).}
\begin{center}\small
\begin{tabular}{lrrrr}
\toprule
Subset & Hit\% & vsMax & RkGrp & MedRk \\
\midrule
painting           & 3.3 & $+1.50$ & 1.44 & 71 \\
painter            & 1.1 & $+1.68$ & 1.30 & 66 \\
first\_name         & 6.7 & $+1.46$ & 1.41 & 90 \\
painter+first\_name & 1.1 & $+1.69$ & 1.30 & 65 \\
all 3              & 3.3 & $+1.55$ & 1.32 & 66 \\
\bottomrule
\end{tabular}
\end{center}

Paintings is the weakest of the four domains. The answer field is the painter's first name, a coarse token with many non-painting associations (\texttt{Claude}, \texttt{Pablo}, \texttt{Leonardo}); and the \texttt{painter} supernode overlaps with \texttt{first\_name} by construction (the painter supernode name already contains the first name). The less-is-more effect is correspondingly weak: the $+3.4$~pp main-text gap is the smallest of the four domains.

\subsection{Cross-domain single-field aggregates}

Averaging over all four domains, the three field types produce very different redirection strengths. \Cref{tab:field-aggregate} reports the aggregate numbers: the \emph{intermediate} field carries the strongest single-field signal ($28.5\%$ Hit), followed by the \emph{answer} field ($24.7\%$), with the \emph{input} field trailing at $4.8\%$. Input-field features encode what the model reads; intermediate- and answer-field features encode what the model should produce.

\begin{table}[h]
\centering\small
\begin{tabular}{lrrrr}
\toprule
Field type   & Hit\% & MedRk & vsMax & RkGrp \\
\midrule
Input        &  4.8 & 131 & $+1.00$ & 2.94 \\
Intermediate & 28.5 &  53 & $+3.34$ & 1.65 \\
Answer       & 24.7 &  33 & $+2.22$ & 1.93 \\
\bottomrule
\end{tabular}
\caption{Per-field aggregates across all four domains. Each row averages the corresponding single-field variant across the four domains, weighted equally.}
\label{tab:field-aggregate}
\end{table}

\section{Adaptive $M$-Search}
\label{appx:msearch}

\subsection{Two-phase protocol}

When a swap pair misses at the default $M_{\mathrm{amplify}}{=}20$, we search for a better amplifier in two phases.

\emph{Phase 1 (coarse geometric probe).} We evaluate $M \in \{0.1, 0.25, 0.6, 1.5, 4.0, 10.0, 20.0\}$ in that order and stop at the first hit. The grid is geometric so that the same number of probes covers both the fine-grained low-$M$ region where some pairs find a hit near $M{\sim}2$ and the $M\approx 20$ neighborhood that the default already explored.

\emph{Phase 2 (KL-transition binary refinement).} If Phase~1 produces no hit, we compute the KL divergence between the steered and unsteered output distributions at each Phase-1 grid point. The KL curve is close to linear in $M$ for most pairs ($R^2 > 0.93$ for 8 of 10 probed USA pairs), but the hits that do exist tend to cluster near the \emph{onset} of meaningful response. We therefore identify the interval in which KL rises most steeply (the ``KL transition''), place a binary-refinement search inside that interval, and run 6 additional probes, accepting a hit at any refinement point. The rationale is that the model's response to multi-feature amplification typically has a sharp onset at a pair-specific $M$, and hits live near that onset.

\subsection{Per-domain rescue totals}

\Cref{tab:msearch-rescue} summarizes the rescue rates split by condition (Labeled, Field-Additivity, and matched-random) and domain.

\begin{table}[h]
\centering\small
\begin{tabular}{llrrrrrr}
\toprule
Cond.\ & Domain & Eligible & New hits & Hit\% & Before & After & $\Delta$ \\
\midrule
Lab.\ & USA       & 1{,}681 & 287 & 17.1 & 31.4 & 43.1 & $+11.7$ \\
Lab.\ & Books     &    84   &   8 &  9.5 &  6.7 & 15.6 &  $+8.9$ \\
Lab.\ & Paintings &    86   &   6 &  7.0 &  4.4 & 11.1 &  $+6.7$ \\
Lab.\ & Products  &   101   &   3 &  3.0 & 23.5 & 25.8 &  $+2.3$ \\
\midrule
FA    & USA       &   740   &  74 & 10.0 & 69.8 & 72.8 &  $+3.0$ \\
FA    & Books     &    21   &   1 &  4.8 & 76.7 & 77.8 &  $+1.1$ \\
FA    & Paintings &    78   &   5 &  6.4 & 13.3 & 18.9 &  $+5.6$ \\
FA    & Products  &    77   &   0 &  0.0 & 41.7 & 41.7 &  $+0.0$ \\
\midrule
Rand.\ & USA       & 2{,}445 & 12 & 0.49 & 0.20 & 0.69 & $+0.49$ \\
Rand.\ & Books     &    90   &  0 & 0.00 & 0.00 & 0.00 & $+0.00$ \\
Rand.\ & Paintings &    90   &  1 & 1.11 & 0.00 & 1.11 & $+1.11$ \\
Rand.\ & Products  &   130   &  8 & 6.15 & 1.52 & 7.58 & $+6.06$ \\
\midrule
\textbf{Total} & --- & \textbf{5{,}623} & \textbf{405} & --- & --- & --- & --- \\
\bottomrule
\end{tabular}
\caption{Adaptive $M$-search rescue rates. ``Eligible'' is the number of pairs that missed at the default ($M_{\mathrm{amplify}}{=}20$, all-three-fields). ``Before'' and ``After'' are the domain-level Hit\% before and after the search. For the labeled (\textsc{Lab.}) and field-additivity (\textsc{FA}) blocks the rescue is per intervention configuration; for the matched-random (\textsc{Rand.\,}) block we apply the same outer adaptive-$M$ harness to all three replicates per labeled pair and score the per-pair best-of-(replicate $\times$ \{default, $M$-tuned\}) under the same lexicographic rule. Labeled sweeps recover between $2.3$ and $11.7$~pp per domain; field-additivity sweeps recover more modestly; the symmetric matched-random sweep recovers $\leq 6.1$~pp in every domain (Products' $+6.1$~pp is the largest, all from a small set of low-$M$ rescues), confirming that the labeled--random gap of \cref{tab:headline} is not an $M$-tuning artefact.}
\label{tab:msearch-rescue}
\end{table}

The overall pattern: $93\%$ of adaptive hits are found in Phase~1 (coarse geometric), and the winning-$M$ distribution is bimodal at $M\sim 2.4$ (47\% of labeled hits) and $M\sim 6.9$ (40\%), with a long tail to $M \geq 10$ (13\%).

\subsection{High-$M$ rescue is null}

To rule out the alternative explanation that the $M{=}20$ default is \emph{too small}, we ran $M \in \{50, 100, 200\}$ on 80 near-miss pairs across three domains. The result was $0$ new hits. Existing hits at $M{=}20$ degraded monotonically as $M$ increased: at $M{=}200$ the generated continuation is typically dominated by a single high-frequency token from the amplification set, with no remaining structural coherence. KL between steered and unsteered saturates above $M\sim 50$, so the model's output distribution is essentially maximally disrupted long before $M$ reaches $200$.

\subsection{Top-$k$ rescue is null}

A second alternative is that the $M{=}20$ signal is distributed over too many features, and that concentrating amplification on the top-$k$ features by graph influence might yield targeted hits at high $M$. On \texttt{products$\to$facebook} we ran the full factorial $k \in \{1, 3, 5, 10, 67\}$ with $M \in \{20, 50, 100, 200\}$ (six source entities, so 120 runs per cell). The result was $0$ hits for any $k < 67$ at any $M$. Only the full 67-feature set achieves hits. A worked example: at $k{=}1$, $M{=}20$ the top-influence feature for \texttt{alibaba$\to$facebook} generates ``the founder of Alibaba in 1999\ldots'', at $k{=}3$ the output is garbled token sequences, at $k{=}10$ the first token is a programming-code token, and only at $k{=}67$ does the model correctly emit ``Mark Zuckerberg\ldots''. Graph influence correlates \emph{negatively} with stored activation in this domain ($\rho{=}-0.362$, $p{=}0.003$): the top-$k$ by graph influence are the loudest but not the most informative features, and concentrating amplification on them produces generic ``question answering'' output rather than the target answer.

\subsection{Five failure modes surfaced by the $M$-sweep}

The pairs that remain unrescued after adaptive search fall into five qualitatively different failure modes. Identifying the five modes was itself a research output: the labels below are pointers to the investigation.

\emph{(1) Severe overshoot.} A pair whose hit exists at small $M$ but is destroyed at the default $M{=}20$. Rescuable by $M$ reduction. Example: \texttt{oklahoma\_tulsa} high-vsMax pairs, $37/42$ pairs rescued at $M{=}5$ (88\% rescue rate). At $M{=}20$ the first token is typically a garbage token (\texttt{AddTagHelper}) followed by ``Oklahoma, is Tulsa'' (the city, not the capital); at $M{=}5$ the first token is a comma and the continuation corrects to ``Oklahoma City, and Tulsa is Oklahoma''.

\emph{(2) Field interference.} A pair that hits on the intermediate+answer subset but not on the all-three-fields variant. Rescuable by field-additivity search (the field-composition effect of \S\ref{sec:results:diagnostic:fields}).

\emph{(3) Feature-interaction noise.} The label is correct but at high $M$ distributed noise masks the target. Example: \texttt{indiana$\to$arkansas} produces \texttt{tonode} at $M{=}20$ and correctly produces \texttt{Little Rock} at $M{=}5$. A subset of these are rescued by $M$ reduction; the remainder are irreducible.

\emph{(4) Signal collapse.} The target logit collapses below a pair-specific $M$ threshold. Example: \texttt{vermont\_burlington$\to$kansas\_wichita} has vsMax $8.34$ at $M{=}20$ but only $0.31$ at $M{=}5$. For these pairs the useful $M$ range is narrow and lies above the default; no value of $M$ in $\{5, 7, 10, 20\}$ produces a hit.

\emph{(5) Feature specificity failure.} The intervention correctly activates the answer concept but the wrong specific entity within the concept. Example: \texttt{vermont$\to$kansas} generates ``Kansas is Hutchinson'' (wrong Kansas city), not ``Topeka''. Not fixable by $M$ adjustment: the features that were grouped under the Kansas supernode appear to encode ``Kansas cities'' rather than ``Topeka'' specifically.

The coarse taxonomy is important for interpreting the aggregate recovery numbers. Modes (1)--(3) are rescued by adaptive $M$-search; mode (4) is occasionally rescued; mode (5) is not.



\section{KL Divergence as a Steering Diagnostic}
\label{appx:kl}

\paragraph{Linearity of KL in $M$.}
Within the $M \in [5, 20]$ range, the per-pair KL curve between the steered and unsteered distributions is close to linear: $R^2 > 0.93$ for 8 of the 10 USA pairs on which we estimated the linear fit. \Cref{tab:kl-pairs} lists the slope and intercept of the linear fit for each probed pair; the slope varies by a factor of $5\times$ across pairs, and is strongly predicted by the pair's ablate count (more source features ablated $\Rightarrow$ steeper KL rise per unit $M$).

\begin{table}[h]
\centering\small
\begin{tabular}{lrrrrr}
\toprule
Pair (source $\to$ target) & Slope $a$ & Intercept $b$ & $R^2$ & amp.\ & total \\
\midrule
kansas $\to$ oklahoma       & 0.318 &  8.28 & 0.941 & 73 & 163 \\
delaware $\to$ oklahoma     & 0.269 &  7.49 & 0.995 & 73 & 152 \\
texas $\to$ oklahoma        & 0.218 &  9.34 & 0.986 & 73 & 138 \\
florida $\to$ oklahoma      & 0.263 & 10.06 & 1.000 & 73 & 139 \\
vermont $\to$ kansas        & 0.079 &  8.43 & 0.864 & 90 & 130 \\
rhode\_island $\to$ wisconsin & 0.194 &  8.16 & 0.982 & 86 & 156 \\
iowa $\to$ utah             & 0.245 &  6.99 & 0.935 & 93 & 253 \\
indiana $\to$ arkansas      & 0.374 &  5.09 & 0.967 & 69 & 269 \\
indiana $\to$ minnesota     & 0.317 &  6.16 & 0.946 & 82 & 282 \\
hawaii $\to$ oklahoma       & 0.109 & 13.69 & 0.848 & 73 & 158 \\
\bottomrule
\end{tabular}
\caption{Per-pair linear fits of KL$(\mathrm{baseline} \parallel \mathrm{steered})$ as a function of $M$, estimated from $M \in \{5, 10, 20\}$. \texttt{amp.} and \texttt{total} count the amplify and total features.}
\label{tab:kl-pairs}
\end{table}

The slope--count correlations are $r(\mathrm{slope}, \mathrm{ablate\_count}) = +0.68$ and $r(\mathrm{intercept}, \mathrm{ablate\_count}) = -0.63$: high-ablate-count pairs have steeper KL rises and lower starting KL, so they are the ones most sensitive to $M$ and most likely to benefit from $M$-reduction.

\paragraph{KL $\geq 12$ as a hit veto.}
Across 30 USA observations, no hit occurs at KL $\geq 12$. As a binary classifier on ``hit iff KL $< 12$'', recall is $7/7 = 100\%$ and precision is $7/19 = 37\%$; the 12 false positives break down into three feature-specificity failures, three evaluator gaps (correct output miscounted due to tokenization differences such as ``St.~Paul'' vs ``Saint Paul''), three signal-collapse pairs, and three near-threshold misses. KL therefore functions as a reliable veto: it reliably rules out failure modes where the output distribution is too disrupted for any target to survive.

\paragraph{Use in adaptive search.}
The adaptive harness uses this in four steps: (i) run at $M{=}20$ and measure KL at position~0; (ii) if KL $\geq 12$, fit the linear KL$(M)$ from one additional probe and compute a pair-specific $M_{\mathrm{crit}} = (12 - b)/a$; (iii) re-run at $M = \lfloor 0.8 \cdot M_{\mathrm{crit}} \rfloor$ with a safety margin; (iv) if the intercept $b$ already exceeds $12$ (as for \texttt{hawaii $\to$ oklahoma}, where $b{=}13.69$), flag the pair as intrinsically disruptive and do not probe further.

\paragraph{A note on target recovery as a signal.}
An earlier version of the analysis promoted \emph{target recovery rate}---whether the target's logit exceeds its own unsteered baseline at any trajectory position---as a primary label-evidence metric, on the basis of a $92\%$ (labeled) vs $29\%$ (random) gap in USA regime~C. A deliberate replication attempt on Books collapsed this gap to $92\%$ vs $89\%$ in the same regime, because the Books model's smaller answer set and distinctive author signatures make random recovery mechanically easy. The lesson, which we took seriously when finalizing the main-paper metrics, is that high-discrimination binary flags may reflect a single domain's structure; \emph{continuous} magnitude metrics (max excess over baseline, vsMax) are more portable. We therefore use Target Recovery only as one of three supporting signals in the within-regime-C analysis of \Cref{sec:results:primary}.


\section{Logit-Shift Regime Taxonomy}
\label{appx:regime}

\subsection{Definitions (position 0)}
We classify each swap by what happens to the target and source logits at position~0, relative to their unsteered baselines. Four regimes matter for interpreting results: \textbf{A} = target up, source down, flip (clean redirection); \textbf{C} = both down, flip (differential disruption---target recovers more than source); \textbf{D} = both down, no flip (generic disruption); \textbf{E} = target flat, source down (pure suppression, no target promotion).

\subsection{Prevalence at the default and at the best variant}
\Cref{tab:regime-prev} compares regime prevalence across three conditions: the per-pair best field-additivity variant, the full labeled ($M{=}20$, all three fields), and the matched-random control under the symmetric per-pair best-of-(replicate $\times$ adaptive-$M$) rule of \cref{tab:headline}. The clean-redirection regime~A is the one we ultimately want to maximize; the generic-disruption regime~D is the one matched-random concentrates in.

\begin{table}[h]
\centering\small
\begin{tabular}{lrrr|rrr}
\toprule
Domain    & \multicolumn{3}{c|}{Regime A (\%)} & \multicolumn{3}{c}{Regime D (\%)} \\
          & best FA & full lab & rand.+$M$-srch & best FA & full lab & rand.+$M$-srch \\
\midrule
USA       & 34.9 &  8.9 & 15.3 &  9.1 & 19.4 & 38.7 \\
Books     & 62.1 & 38.8 & 26.7 &  3.3 &  3.3 & 34.4 \\
Products  & 62.1 & 56.8 & 28.8 &  2.3 &  2.3 & 18.9 \\
Paintings & 47.8 & 17.8 & 16.7 &  2.2 &  6.7 & 36.7 \\
\bottomrule
\end{tabular}
\caption{Regime A (clean redirection) and Regime D (generic disruption) prevalence under the per-pair best field-additivity variant, full labeled, and the symmetric per-pair best-of-(replicate $\times$ adaptive-$M$) matched-random condition. The best variant dramatically amplifies regime A and suppresses regime D compared to full labeled; the matched-random control concentrates in regime~D ($35$--$39\%$ in USA, Books, and Paintings) even with the symmetric $M$-search harness.}
\label{tab:regime-prev}
\end{table}

Three patterns stand out. The best FA variant pushes many more cases into regime~A (clean redirection) than the full labeled variant does (USA: $8.9 \to 34.9\%$; Books: $38.8 \to 62.1\%$); removing input-field features eliminates the generic disruption that was pushing cases into regime~C/D.

Within regime~A, hit rates also improve: USA goes from $44\%$ (full labeled) to $80\%$ (best variant); Books from $4.3\%$ to $54.4\%$. The best variant is not merely redistributing pairs across regimes---it is also raising the quality of each regime.

Regime~D (generic disruption, no flip) nearly vanishes under the labeled condition ($19.4 \to 9.1\%$ in USA), while the symmetric matched-random+$M$-srch control concentrates there ($35$--$39\%$ in USA, Books, and Paintings; $19\%$ in Products). This is why vsMax separates labeled from random even when both conditions produce high suppression rates and even when the random side is given the same adaptive-$M$ harness.

\subsection{Within-regime-C signals (USA)}
Regime~C is the largest and most ambiguous regime in the labeled condition: both logits drop, but the target less so. The richest within-regime-C signals are given in \Cref{tab:regime-c-usa}, which is the data behind the within-regime-C result reported in \Cref{sec:results:primary}.

\begin{table}[h]
\centering\small
\begin{tabular}{lrr}
\toprule
Signal & Labeled & Random \\
\midrule
Target-recovery rate (\%)      & 92.2 & 29.3 \\
Sustained dominance (\texttt{tgt\_win\_pct}) & 0.673 & 0.319 \\
Mean vsMax                     & $+2.33$ & $-0.10$ \\
Hit\% (conditional on regime C) & 24.8 & 1.8 \\
\bottomrule
\end{tabular}
\caption{Within-regime-C signals, USA. Labeled features cause differential disruption in which the target recovers above its own unsteered baseline while the source stays suppressed; matched-random does not.}
\label{tab:regime-c-usa}
\end{table}

The target-recovery signal (whether the target's logit at any point exceeds its own unsteered baseline) is strong in USA regime~C but does not generalise: in Books regime~C the labeled vs.\ random gap collapses to near zero. Books has a small answer space and sharply distinct entity signatures, so any large enough perturbation mechanically pushes the target above its baseline. Target-recovery therefore serves as one of three supporting within-regime signals.


\section{Natural Cluster Geometry of CPAS Metrics}
\label{appx:clustering}

The decision tree (\cref{tab:decision-thresholds}) partitions features into four functional roles using six axis-aligned threshold cuts. A natural concern is whether these four labels reflect dense regions of the feature metric space or impose arbitrary cuts. We tested it by re-aggregating per-feature metrics across all four datasets ($\approx 8{,}000$ deduplicated features) and running unsupervised clustering with several methods.

\paragraph{Threshold sensitivity.}
Each of the four primary cuts was perturbed by $\pm 10\%$ on the manifest. Per-cut flip rates ranged from $0.05\%$ to $1.5\%$, and within tight neighborhoods of each threshold the local flip rate stayed below $4\%$. The simplest ``many points sit on arbitrary boundaries'' reading is therefore inconsistent with the data.

\begin{figure}[h]
  \centering
  \includegraphics[width=\textwidth]{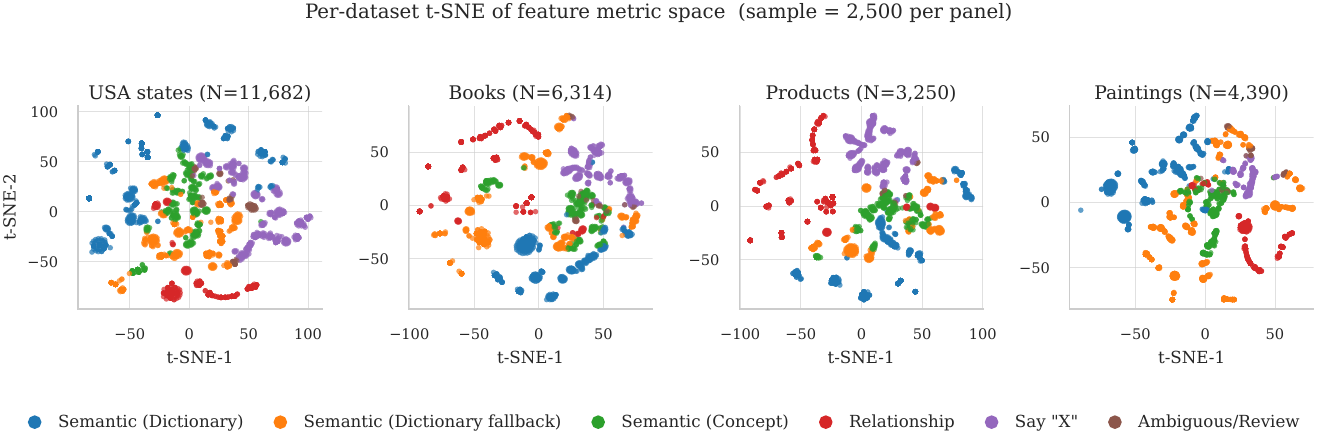}
  \caption{Per-dataset t-SNE of the standardized 6-metric CPAS vector. Points are coloured by rule label. \textsc{Relationship} and \textsc{Say-X} form tight clusters; \textsc{Sem-Dict} is multi-modal, an early hint of the bimodality discussed below.}
  \label{fig:tsne-per-dataset}
\end{figure}

\paragraph{Visual structure.}
We standardized the six metrics and projected them with t-SNE per dataset (\cref{fig:tsne-per-dataset}). Across all four domains the four rule labels occupy spatially coherent regions: \textsc{Relationship} forms a single low-sparsity arc, \textsc{Say-X} a tight late-layer cluster, and \textsc{Sem-Conc} occupies the layer-shallow midband. The \textsc{Sem-Dict} bucket consistently splits into two visually disconnected regions, an early hint that the rule's semantic-fallback branch sweeps up two genuinely different feature populations.

\begin{figure}[h]
  \centering
  \includegraphics[width=\textwidth]{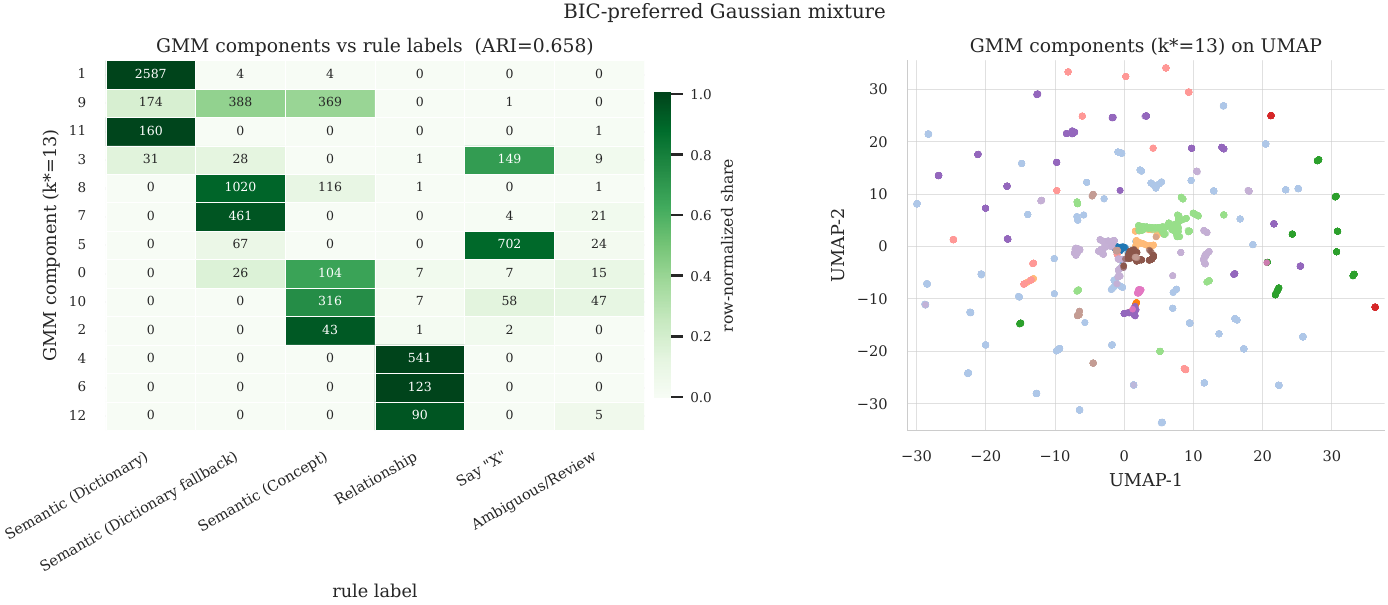}
  \caption{Left: Gaussian-mixture BIC as a function of the number of components for full and diagonal covariance, on the deduplicated manifest. Both curves prefer $k\in\{10,12\}$. Right: per-dataset UMAP coloured by the BIC-preferred GMM partition; the rule labels (overlaid markers) are consistent with the natural geometry but coarser than it.}
  \label{fig:gmm-bic}
\end{figure}

\paragraph{Natural number of components.}
Three principled criteria for the natural number of clusters disagree with the rule's $k{=}4$. Density-based clustering (HDBSCAN, \texttt{min\_cluster\_size}$=150$) settles at $13$ components after excluding noise; Ward agglomerative silhouette plateaus at $k{=}8$; and Gaussian-mixture BIC monotonically prefers $k{=}10$ (full covariance) with a BIC drop of $1.13\times 10^5$ nats from $k{=}4$, far beyond any plausible noise level. \cref{fig:gmm-bic} shows the BIC curve and a per-dataset UMAP overlay of the BIC-preferred clusters.

\paragraph{Coarse-grained recovery.}
Despite the $k{>}4$ preference, the rule labels are an Adjusted-Rand-Index--coherent coarse-graining of the natural geometry: KMeans-$k{=}4$ yields ARI $\approx 0.51$ against the rules, and the BIC-preferred $k{\approx}10$ yields ARI $\approx 0.62$. Per-dataset transfer of the rule geometry varies (\cref{tab:cluster-ari}): the rule partition transfers cleanly to Products and Books, weakly to Paintings, and only partially to USA, consistent with USA's heavier mix of \textsc{Sem-Dict}-fallback features and Paintings' smaller, lower-quality circuit population (\cref{appx:negative}).

\begin{table}[h]
\centering\small
\begin{tabular}{lrr}
\toprule
Dataset & KMeans $k{=}4$ ARI & GMM diag $k{=}4$ ARI \\
\midrule
USA       & 0.35 & 0.35 \\
Books     & 0.50 & 0.51 \\
Products  & 0.60 & 0.62 \\
Paintings & 0.24 & 0.25 \\
\bottomrule
\end{tabular}
\caption{Per-dataset Adjusted Rand Index between rule labels and unsupervised $k{=}4$ partitions of the standardized 6-metric vector.}
\label{tab:cluster-ari}
\end{table}

The four-role partition is best understood as a deliberately coarse-grained labeling of a finer natural manifold. The rule labels are spatially coherent, threshold-stable, and ARI-coherent at $k{=}4$, but the data has more structure than four classes: the \textsc{Sem-Dict} fallback in particular is bimodal and is a natural target for a refined taxonomy in future work (\cref{sec:discussion}).

\section{Human-curated Dallas Case Study}
\label{appx:dallas_case}

The Dallas/Austin attribution graph is the only target circuit in our datasets that has a publicly-released human-curated subgraph (the Neuronpedia graph \texttt{gemma-fact-dallas-austin} by user \texttt{mh2parker}). We use this single-circuit asymmetry to run a focused case study: hold the target circuit fixed at \texttt{texas\_dallas}, run every non-Dallas USA state (49 sources) as the source side, and vary the target-side feature \emph{bag} across four conditions: human curation, our auto pipeline, an unlabeled top-$K$-by-influence control, and a label-shuffled floor control. The headline comparison is ours-vs-human; the top-$K$ family is a control that asks whether pure influence ranking can substitute for either, and the shuffled-labels condition is a sanity floor.

\subsection{Setup}
One target circuit (\texttt{texas\_dallas}, the Dallas/Austin prompt with $1{,}182$ features at our cumulative-influence threshold), 49 non-Dallas USA source states (every entity from \texttt{full\_swap\_human\_dallas.yml}), Dallas always the target. Each \emph{condition} is a different target-side bag for the Dallas circuit:

\begin{itemize}\itemsep0pt
\item \textbf{ours}: full canonical auto Dallas grouping (the eight concept-aligned supernodes of \cref{fig:circuit-texas}, $458$ classified features). Per-pair best of the 7-variant field-additivity sweep with adaptive $M$-search.
\item \textbf{human}: the $22$ features pinned in the Anthropic Neuronpedia public graph \texttt{gemma-fact-dallas-austin} (5 named supernodes; manifest at \\ \texttt{output/usa\_states\_fact\_batch/\_swap\_conditions/} \\ \texttt{human\_dallas/texas\_dallas/manifest.json}). Same field-additivity + $M$-search protocol.
\item \textbf{top-$K$ (control, single-bag)}: top-$K$ Dallas features by max \texttt{node\_influence} from the canonical $1{,}182$-feature universe, with $K \in \{10, 21, 100, 200\}$, run as a flat unlabelled bag --- \emph{no} field-additivity, \emph{no} concept-field semantics, \emph{no} \texttt{supernode\_name} filter, \emph{no} per-variant subsetting. Adaptive $M$-search enabled. This is the fair influence-only control: an earlier version of the experiment ran the top-$K$ family with \texttt{control.mode: additivity} and inherited \texttt{auto}'s supernode labels onto the top-$K$ rows, which let the field-additivity matcher subset the bag per-variant; that variant of the protocol is no longer used (see comparison below).
\item \textbf{shuffled-labels (floor control)}: same $22$ features as human, supernode labels permuted; 7 field-additivity variants + $M$-search.
\end{itemize}

The source side (the 49 source states' ablations) is the canonical auto Dallas-target grouping in every condition --- on average $1{,}253$ features per state --- so the only quantity that varies across rows is the \emph{target bag}. The source-side bag is not held to any matching constraint, just held \emph{constant} across conditions (asymmetric design). Hit metric is as defined in \S\ref{appx:specificity}; for this target ``Austin'' is a single six-letter word, so rule~(iii) detects the same hits as a first-subword check in nearly all cases. Source hit-rate is the fraction of the 49 source states for which at least one cell hits.

\subsection{Headline result: ours $\approx$ human $\gg$ top-$K$, shuffled = $0$}
\Cref{tab:dallas-case-headline} gives the per-condition source coverage and per-cell efficiency. Both labeled methods (ours, human) clear $78\%$ of source states; every fair single-bag top-$K$ saturates below $13\%$; shuffled-labels is at $0\%$. \emph{The labeled methods reach the same source-coverage ceiling, but at very different per-cell costs}: human uses $4$ amplified features per call on average and median $M_{\mathrm{tuned}}{=}2.0$ (collapsed near default), our auto pipeline uses $96.5$ features per call and median $M_{\mathrm{tuned}}{=}4.1$ (consistent across pairs).

\begin{table}[h]
\centering\small
\begin{tabular}{lrrrr r}
\toprule
Condition & Sources hit / 49 & Cell hit-rate & Mean amp.\ feat / call & $M_{\mathrm{tuned}}$ median & cum infl / call \\
\midrule
ours      & 40 / 49 (82\%)   & 30.1\%        & 96.5                   & 4.08 & 0.067 \\
human     & 38 / 49 (78\%)   & 26.2\%        & 4.0                    & 2.00 & 0.014 \\
top-100   &  6 / 49 (12\%)   & 12.2\%        & 100.0                  & 2.40 & 0.146 \\
top-21    &  3 / 49 (6\%)    & 6.1\%         & 21.0                   & 6.93 & 0.073 \\
top-10    &  3 / 49 (6\%)    & 6.1\%         & 10.0                   & 20.0$^{\dagger}$ & 0.051 \\
top-200   &  1 / 49 (2\%)    & 2.0\%         & 200.0                  & 4.08 & 0.197 \\
shuffled-labels & 0 / 49 (0\%) & 0.0\%       & 3.4                    & ---  & 0.004 \\
\bottomrule
\end{tabular}
\caption{Per-condition source coverage on the 49-state Dallas-target swap (target = \texttt{texas\_dallas}; 49 non-Dallas USA states as sources). \emph{Cum infl / call} is the mean per-pair cumulative \texttt{node\_influence} consumed by the target-side amplification bag --- the $x$-axis of \cref{fig:topk-saturation-full50}. Top-$K$ rows are the fair single-bag re-run; ours/human/shuffled use the standard 7-variant field-additivity sweep. $^{\dagger}$top-10 hits all occur at the default $M{=}20$; no $M$-search refinement was needed.}
\label{tab:dallas-case-headline}
\end{table}

\Cref{fig:topk-saturation-full50} plots successful swaps (\%) (y) against mean cumulative influence per swap (x). The three green top-$K$ markers (top-21, top-100, top-200; top-10 omitted from the figure as redundant with top-21) sit near the floor; \texttt{top-21} sits just above ours ($1.09\times$ cumulative influence), \texttt{top-100} at $2.17\times$, and \texttt{top-200} at $2.93\times$. None clears the $\sim 12\%$ successful-swaps ceiling. The labeled markers (red, blue) sit at $\sim 80\%$. The non-monotone $K{=}100 \to K{=}200$ drop ($12\% \to 2\%$) is consistent with high-$K$ disruption (every $K{=}200$ hit needed $M{=}4.1$ via $M$-search; the default $M{=}20$ never hit).

\subsection{Per-source heatmap: methods are non-redundant}
The source-coverage ceiling hides genuine method-specific wins. Of the 49 sources:

\begin{itemize}\itemsep0pt
\item $22 / 49$ are universally easy (every method except shuffled hits).
\item $1 / 49$ is universally hard (\texttt{idaho\_idaho\_falls}; no method hits).
\item $26 / 49$ form a disagreement set in which ours and human partially complement each other, and a small number of sources are accessible only to top-$K$:
  \begin{itemize}\itemsep0pt
  \item \emph{human-only}: \texttt{missouri\_kansas\_city}, \texttt{north\_dakota\_fargo}. The 22-feature human curation captures something neither ours nor any of the four top-$K$ controls reaches.
  \item \emph{ours-only}: \texttt{colorado\_colorado\_springs}. The label-driven supernode composition captures something even top-200 misses.
  \item \emph{top-200-only}: \texttt{oklahoma\_tulsa} (in the previous unfair phase3v3 protocol). A pure-volume influence win that disappears in the fair single-bag re-run; the source is hit only when the field-additivity boost is also active.
  \end{itemize}
\end{itemize}

The two human-only sources are noteworthy: a $22$-feature curated bag finds a redirection that an automated pipeline with $\sim 100$ features and influence ranking miss. Our auto pipeline reciprocally finds at least one source neither curation nor any single-bag top-$K$ reaches.

\subsection{Naturalness: $M_{\mathrm{tuned}}$ distributions}
Across all hits, $M_{\mathrm{tuned}}$ medians sit at $2.0$ (human), $4.1$ (ours), and $2.4$ (top-100); only \texttt{top-21} required $M{\sim}6.9$ on its three M-search hits. Human is the most \emph{natural}: when it hits, it hits at small $M$, and many of its hits already exist at the default. Ours is the most \emph{consistent}: tight $M_{\mathrm{tuned}}$ IQR, no high-$M$ outliers above ${\sim}7$, and all 40 hits cluster within a $4$-point $M$ window. The high-$K$ controls drift toward smaller $M$ (top-100 at $M{=}2.4$, top-200 at $M{=}4.1$) because the default $M{=}20$ overshoots when $K$ is large --- $M$-search is doing all the work for these conditions.

\subsection{Top-$K$ saturation as a control: precise protocol}
The fair top-$K$ family is included to test whether pure influence-only ranking can substitute for either the human curation or our pipeline. Each top-$K$ condition runs the following protocol:

\begin{itemize}\itemsep0pt
\item Target bag: top-$K$ Dallas features by max \texttt{node\_influence} over the canonical Dallas grouping universe. Source bag: canonical auto source's full grouping ($\sim 1{,}253$ features per state on average, no filter, no field-additivity).
\item No \texttt{concept\_fields}, no \texttt{supernode\_name} filtering, no per-variant subsetting --- one intervention per pair (\texttt{control.mode: single\_bag\_grouping}).
\item Adaptive $M$-search: Phase 1 coarse $M \in \{0.1, 0.25, 0.6, 1.5, 4.0, 10.0, 20.0\}$; Phase 2 KL-binary refinement; same parameters as \texttt{fullscale\_usa\_labeled\_msearch.yml}.
\item Hit metric: as defined in \S\ref{appx:specificity} (``Austin'' as a six-letter word matches via rule~(iii)). Source hit-rate: fraction of the 49 source states with at least one hit.
\item Configs: \texttt{phase4\_topk\_\{10,21,100,200\}\_dallas\_singlebag.yml}; control: \texttt{scripts/experiments/batch/pipeline/controls/single\_bag\_grouping.py}; launcher: \texttt{tools/launch\_phase4\_topk\_singlebag.sh}; aggregator: \texttt{tools/phase4\_topk\_singlebag\_aggregate.py}.
\end{itemize}

The top-$K$ saturation was originally measured (Phase B v3, May 5) with \texttt{control.mode: additivity}, which transferred \texttt{auto}'s supernode labels onto the top-$K$ rows and let the field-additivity matcher subset the bag per-variant. That earlier version reported top-21 / top-100 / top-200 source coverage of $63\% / 69\% / 82\%$, suggesting top-$K$ saturated near \texttt{ours}. The fair re-run with the same target bags but no field-additivity gives $6\% / 12\% / 2\%$. The $57$--$80$~pp gap was the field-additivity boost masquerading as influence-ranking quality. The result reported in this section is the fair version; the cross-domain influence-matched analogue (\S\ref{appx:topkim}) reaches the same conclusion on four in-scope domains under a per-pair budget-match.

\subsection{Threats to validity}
Single target circuit (Dallas) and single domain (USA states); the Dallas top-$K$ also happens to be unusually concept-pure ($\sim 90\%$ of top-$21$ features fall in the eight \texttt{auto}-labeled supernodes), so the field-additivity boost reported above may be larger here than on a domain with messier labels. The cross-domain influence-matched experiment (\S\ref{appx:topkim}) runs the analogous test on four domains under a stricter per-pair budget-match.

The source-side ablation uses canonical auto's labels for every condition, so the auto pipeline is ``running'' on every row as ablation, even when the target side is human/top-$K$/shuffled. This is the standard convention for this experiment (the only Dallas-target curated graph is the human one, so we cannot run a symmetric human-on-both-sides bag), but it means we cannot disentangle ``labels matter for ablation'' from ``labels matter for amplification'' in this specific case study. \S\ref{appx:topkim} runs both sides influence-matched and reaches the same direction of result.


\section{Per-Domain Full-Scale Labeled vs Random}
\label{appx:specificity}

\paragraph{Symmetry of the per-pair best-of construction.}
The labeled FA+$M$-srch column of \cref{tab:headline} reports the per-pair best across $\{\textsc{field-additivity variants}\} \times \{\text{default}, M\text{-tuned}\}$. To make the matched-random control comparable we apply the \emph{same} per-pair best-of rule to its three replicates: for every labeled pair we take the best across $3 \text{ replicates} \times \{\text{default}, M\text{-tuned}\}$ random candidates under the same lexicographic score (hit, then $-$rank, then vsMax). This means both columns get one row per labeled pair, both columns see the same outer adaptive-$M$ harness, and both columns are scored under the unified hit rule below. The per-replicate random row at the default $M$ is kept in \cref{tab:fullscale-spec} as a baseline for the rescue accounting in \cref{tab:msearch-rescue}, but it is not the right object to compare against the labeled FA+$M$-srch number; that comparison happens between \cref{tab:headline}'s \emph{Rand.\ +$M$-srch} column and the matching per-pair-best random+$M$-srch row of \cref{tab:fullscale-spec}.

\paragraph{What counts as a hit.}
Throughout the paper a swap is a \emph{hit} when the target answer is
detected in the model's steered output by any of three simple text
checks, applied in order: (i)~the full target string appears in the
output (with punctuation and hyphens normalised, so ``St.\ Paul'' and
``Saint Paul'' match); (ii)~the first emitted subword is a substring
of the target answer (so emitting ``Mark'' counts toward ``Mark
Zuckerberg''); (iii)~any content word of the target of length
$\geq 3$ characters appears as a whole word in the output (so
``Harper'' anywhere in the continuation counts toward ``Nelle Harper
Lee''). A small per-domain blacklist removes generic words that would
match too easily (e.g.\ ``city'' on the USA panel). We arrived at
this rule by hand-checking outputs that a stricter first-subword
equality rule scored as misses: most were correct redirections that
emitted the answer one token late, with a different tokenisation, or
surrounded by punctuation; the three checks above recover those cases
while the word-boundary requirement and the length-$3$ minimum keep
stop-words and one-letter tokens out. Every comparison condition
(matched-random, top-$K$ influence-matched, shuffled labels) is
scored under the same rule, so the labeled--control gap is what
carries the claims. A future revision can
swap the three text checks for an LLM-as-judge that grades the
steered output against the target answer with full lexical and
semantic flexibility; the three checks here are a deterministic
placeholder for that judge and the rest of the paper does not
otherwise depend on the choice.

\Cref{tab:fullscale-spec} gives the full per-domain comparison at the standard defaults (all-three-fields, $M_{\mathrm{ablate}}{=}-2$, $M_{\mathrm{amplify}}{=}20$, attention not frozen). Each domain is reported with the labeled intervention, the matched-random-control replicate set at the default $M$ (three replicates per labeled pair, so per-replicate random $N$ is $3\times$ the labeled $N$), and the per-pair best of the matched-random set under the symmetric adaptive $M$-search (one row per labeled pair, taking the best across the three replicates and their $M$-tuned outputs).

\begin{table}[h]
\centering\small
\begin{tabular}{llrrrrrrr}
\toprule
Domain & Cond.\ & $N$ & Hit\% & Supp\% & vsMax & RkGrp & MedRk & Flip\% \\
\midrule
USA       & labeled         & 2{,}450 & 31.4 & 92.8 & $+2.86$ & 1.72 & 5   & 98.2 \\
          & random          & 7{,}350 &  4.6 & 83.4 & $-2.31$ & 9.00 & 566 & 69.2 \\
          & random+$M$-srch & 2{,}450 &  0.7 & 84.1 & $-1.23$ & 6.32 & 148 & 75.5 \\
Books     & labeled         &    90   &  6.7 & 64.4 & $+6.70$ & 1.03 & 17  & 96.7 \\
          & random          &   270   &  0.0 & 88.1 & $-0.73$ & 2.43 & 283 & 79.6 \\
          & random+$M$-srch &    90   &  0.0 & 85.6 & $+0.19$ & 2.18 & 266 & 83.3 \\
Products  & labeled         &   132   & 23.5 & 64.4 & $+3.46$ & 1.20 & 26  & 97.0 \\
          & random          &   396   &  0.5 & 87.4 & $+0.14$ & 2.25 & 354 & 75.3 \\
          & random+$M$-srch &   132   &  7.6 & 84.8 & $+1.17$ & 1.79 & 262 & 71.2 \\
Paintings & labeled         &    90   &  4.4 & 37.8 & $+1.50$ & 1.31 & 70  & 97.8 \\
          & random          &   270   &  0.0 & 74.4 & $+0.12$ & 1.96 & 196 & 88.9 \\
          & random+$M$-srch &    90   &  1.1 & 68.9 & $+1.27$ & 1.42 & 187 & 90.0 \\
\bottomrule
\end{tabular}
\caption{Full-scale labeled vs.\ matched-random comparison. The first two sub-rows per domain use the all-three-fields default at $M_{\mathrm{amplify}}{=}20$ (labeled and matched-random at three replicates per pair). The third sub-row reports the matched-random control under the same per-pair best-of-(replicate $\times$ adaptive-$M$) construction as the labeled FA+$M$-srch column of \cref{tab:headline}, so the comparison is symmetric: every labeled pair is matched to exactly one random-control row, taken as the best of $3 \times \{$default, $M$-tuned$\}$ candidates by the same lexicographic score (hit, then -rank, then vsMax). The labeled--random vsMax gap is the central signal of operational usefulness (\S\ref{sec:results:primary}); applying $M$-search to the random side narrows the per-replicate Hit\% gap by at most $7$\,pp (Products) and leaves the labeled--random ordering unchanged in every domain.}
\label{tab:fullscale-spec}
\end{table}

The rightmost four columns tell the same qualitative story from different angles. The labeled intervention promotes the target to a median rank inside the top $70$ in every domain, while the matched-random controls leave it near rank $200$--$550$. \texttt{Flip\%} (how often the target's logit overtakes the source at any trajectory position) is above $96\%$ in every labeled condition, which is why the operational test relies on vsMax rather than on the flip indicator alone.

\paragraph{Labeled--random vsMax gap by domain.}
Against the per-replicate matched-random control at the default $M$, Books and USA show the largest labeled--random vsMax (logit margin over the next-best answer) gap ($+7.4$ and $+5.2$ respectively); Products is moderate ($+3.3$) and Paintings weak ($+1.4$). The same ordering holds when the random side is replaced by the symmetric per-pair best-of-(replicate $\times$ adaptive-$M$) random+$M$-srch row of \cref{tab:fullscale-spec}: Books $+6.5$, USA $+4.1$, Products $+2.3$, Paintings $+0.2$. This ordering matches the operational-usefulness verdict, which is one reason vsMax is a primary metric.


\section{Per-Pair Influence-Matched Top-$K$ Baseline}
\label{appx:topkim}

The matched-random control fixes the feature count and per-layer histogram but not the total graph influence the intervention uses. A labeled pair whose features happen to land on high-influence nodes has a structural advantage unrelated to concept alignment. This section adds a second control---\textsc{Top-$K$ Influence-Matched}---that instead holds the per-side cumulative \texttt{node\_influence} budget fixed at the labeled best-of value.

\subsection{Construction}

For every pair $(e_A, e_B)$ in the four in-scope domains, let $S^{\mathrm{lab}}$ be the source-side ablated feature set and $T^{\mathrm{lab}}$ the target-side amplified feature set selected by the per-pair best-of-(field-additivity~$\times$~\{default, $M$-search\}) labeled run. Define
\[
B^{\mathrm{src}}(e_A, e_B) = \sum_{f \in S^{\mathrm{lab}}} \mathrm{influence}_{e_A}(f),
\qquad
B^{\mathrm{tgt}}(e_A, e_B) = \sum_{f \in T^{\mathrm{lab}}} \mathrm{influence}_{e_B}(f),
\]
where $\mathrm{influence}_{e}(\cdot)$ is the per-feature \texttt{node\_influence} from $e$'s attribution graph, deduped by max over $(\text{layer}, \text{id})$. The matched control's bag is the smallest top-$K$ prefix of each entity's own influence-ranked grouping universe whose cumulative influence reaches the corresponding budget:
\[
K^{\mathrm{src}} = \min\!\left\{K : \sum_{f \in \text{top-}K(e_A)} \mathrm{influence}_{e_A}(f) \geq B^{\mathrm{src}}\right\},
\]
and analogously $K^{\mathrm{tgt}}$ on the target side. The intervention is a single bag, with no field-additivity sub-selection, with $M_{\mathrm{ablate}}{=}-2$, $M_{\mathrm{amplify}}{=}20$, and the same outer adaptive-$M$ sweep enabled. The grouping universe used for the top-$K$ is the  same set of steerable features the labeled pipeline classifies; scaffold/error rows ($f{=}-1$) and embedding/logit-only rows ($\text{layer}{<}0$) are excluded by construction. 

\subsection{Result}

\Cref{tab:topk-im} reports the head-to-head Hit\% comparison and a paired McNemar test (a paired sign test on the hit/miss contingency) on the contingency ($b{=}{}$labeled-only wins, $c{=}{}$top-$K$-only wins). Labeled features beat the influence-matched top-$K$ baseline in every domain where the test has power: paired $p$-values are $\ll 10^{-30}$ (USA), $2.7\!\times\!10^{-20}$ (Books), $1.1\!\times\!10^{-13}$ (Products), and $0.23$ (Paintings, $N{=}56$, underpowered). The companion 4-panel figure (\cref{fig:topk-im-4domains}) plots each domain's Hit\% gap against the per-condition mean number of amplified features, confirming that the labeled bag uses substantially \emph{more} features than top-$K$ at the same per-pair influence budget, while still scoring higher on the redirection metric.

\begin{table}[h]
\centering
\small
\caption{Per-pair influence-matched top-$K$ baseline vs.\ labeled best-of, four domains. For every swap pair the top-$K$ baseline selects the smallest prefix of features ranked by per-entity graph influence whose cumulative influence matches the labeled best-of budget; both conditions share the same outer $M$-search sweep. Hit\% is scored under the unified rule (\S\ref{appx:specificity}) on the demo cross-run intersection. $b$ = pairs where only labeled hits; $c$ = pairs where only top-$K$ hits (contingency cells for the paired McNemar test). Paintings $N{=}56$ because the influence-matched run covers only $56$ of the $90$ demo-intersection pairs; that subset is underpowered. Full per-pair join in \texttt{output/research/topk\_im\_pairs\_<domain>.csv}.}
\label{tab:topk-im}

\begin{tabular}{lrrrrrrr}
\toprule
& & \multicolumn{2}{c}{Hit\%} & & \multicolumn{2}{c}{McNemar} & \\
\cmidrule(lr){3-4}\cmidrule(lr){6-7}
Domain & $N$ & top-$K$ & Lab.\ & $\Delta$ pp & $b$ & $c$ & $p$ \\
\midrule
USA       & 2{,}450 &  4.2 & 72.8 & $+68.6$ & 1{,}686 &  5 & $\ll\!10^{-30}$ \\
Books     &      90 &  4.4 & 77.8 & $+73.3$ &      66 &  0 & $2.7\!\times\!10^{-20}$ \\
Products  &     132 &  1.1 & 50.0 & $+48.9$ &      44 &  0 & $1.1\!\times\!10^{-13}$ \\
Paintings &      56 &  7.1 & 16.1 & $+8.9$  &       8 &  3 & $0.23$ \\
\bottomrule
\end{tabular}
\end{table}

The labeled bag uses substantially more features than top-$K$ at the same per-pair influence budget. Two readings are consistent with this: (i) a substantial fraction of the labeled features are low-influence ``scaffold'' nodes that nevertheless gate the answer circuit; (ii) the highest-influence nodes are the \emph{loudest} features at the first token, but not the most informative for redirecting the model to a specific target, consistent with \S\ref{appx:msearch}'s top-$k$-rescue null on Products. We do not separate (i) and (ii) within this experiment.


\section{Suppression Is Easy; Steering Is Hard}
\label{appx:suppression}

The most common outcome of a labeled intervention is not a successful redirect but a \emph{suppression without retargeting}: the source answer disappears from the output, but the target does not appear either. This section documents that asymmetry and explains why it is structurally expected.

\subsection{The gap between suppressing and steering}

Suppression and targeting are not symmetric operations. Ablating source features at $M_{\mathrm{ablate}}{=}{-}2$ multiplies each ablated feature's decoder contribution by a negative scalar, driving it to near zero regardless of what that feature encodes semantically; any collection of features that covers the source-entity circuit is sufficient to kill the source token. Amplifying target features only promotes the target if those features actually encode the answer---they must carry the correct answer's identity in their activation. The result is that suppression requires coverage; targeting requires specificity.

\subsection{Suppression rates vs.\ targeting rates}

\Cref{tab:supp-vs-hit} compares source-suppression rates with target-detection rates for the labeled intervention and the matched-random control across all four domains.

\begin{table}[h]
\centering\small
\begin{tabular}{lrrrrrr}
\toprule
 & \multicolumn{2}{c}{Labeled} & \multicolumn{2}{c}{Matched-random} & \multicolumn{2}{c}{Rand.\ +$M$-srch} \\
\cmidrule(lr){2-3}\cmidrule(lr){4-5}\cmidrule(lr){6-7}
Domain & Supp\% & Hit\% & Supp\% & Hit\% & Supp\% & Hit\% \\
\midrule
USA      & 92.8 & 31.4 & 83.4 & 4.6 & 84.1 & 0.7 \\
Books    & 64.4 &  6.7 & 88.1 & 0.0 & 85.6 & 0.0 \\
Products & 64.4 & 23.5 & 87.4 & 0.5 & 84.8 & 7.6 \\
Paintings & 37.8 & 4.4 & 74.4 & 0.0 & 68.9 & 1.1 \\
\bottomrule
\end{tabular}
\caption{Source-suppression rate (Supp\%) and target-detection rate (Hit\%) for labeled, matched-random (per replicate at the default $M$), and the symmetric per-pair best-of-(replicate $\times$ adaptive-$M$) matched-random condition. Random interventions suppress the source at rates comparable to labeled but land almost no hits, and giving the random side the same outer adaptive-$M$ harness as the labeled FA+$M$-srch column does not close that gap (only Products picks up a few hits, $0.5\rightarrow7.6\%$, against $41.7\%$ for FA+$M$-srch on the same pair set).}
\label{tab:supp-vs-hit}
\end{table}

The labeled and matched-random conditions suppress the source at broadly similar rates---within $25$~pp in three of four domains---while their Hit\% differs by a factor of $5$--$20\times$. A random bag of features matched in count and layer distribution achieves $74$--$88\%$ suppression in most domains while hitting essentially zero targets. This confirms the asymmetry: the source disappears because the circuit is disrupted; the target appears only when features that encode the specific answer are present.

\subsection{The suppression-only zone}

Most suppressed pairs land in what we call the \emph{suppression-only zone}: the source answer is absent but the target is not detected, so the steered output fills the slot with generic or incoherent text. In the labeled condition at the all-fields default, the suppression-only zone accounts for $63.3\%$ (USA), $66.2\%$ (Books), $44.3\%$ (Products), and $36.3\%$ (Paintings) of all pairs---the single largest outcome bucket in every domain. These correspond to regime~E (source logit drops, target stays flat) and the subset of regime~C/D pairs that never produce a detectable target token.

The suppression-only zone shrinks substantially when the best field-additivity variant replaces the all-fields default (\S\ref{appx:fieldadd}): removing input-field features that compete with the answer circuit moves many pairs from suppressed-only to a clean hit. This is the mechanism behind the less-is-more effect reported in the main text.

\subsection{Why suppression is structurally cheaper}

Three reasons explain the asymmetry:

\emph{(a) Ablation is semantics-blind.} The $-2\times$ multiplier drives a feature's contribution toward zero independently of what that feature encodes. Any bag large enough to cover the source-entity circuit suppresses the source.

\emph{(b) Random bags are sufficient for suppression.} A matched-random bag---same feature count and per-layer histogram as the labeled bag, but drawn from outside concept-aligned supernodes---suppresses the source at nearly the same rate as labeled features (Supp\% gap $\leq 25$~pp in three of four domains). This is why Supp\% is a poor discriminator between conditions; it is not what separates labeled from random.

\emph{(c) Targeting distributes probability mass over hundreds of answers.} Suppressing the source spreads probability mass across the model's full vocabulary. To install the target answer, the amplified features must collectively shift the target logit above all competitors---a much harder coordination problem than eliminating the incumbent.

\subsection{Implication for evaluation}

Because suppression is almost universal under both labeled and random conditions, Supp\% does not distinguish between a conceptually-driven intervention and a generic disruption. Hit\% and vsMax (the logit margin over the next-best answer) are the metrics that track this distinction, and are therefore the primary evaluation signals throughout the paper.


\section{Feature Stability Across Prompt Phrasings}
\label{appx:crossprompt}

The features assigned to an entity should reflect the concept itself across different prompts. We test this directly: for each entity we generate five differently-worded prompts that all call for the same answer, extract the attribution graph for each, and ask whether the same features appear consistently across all five. We ran this analysis on $1{,}592$ entity pairs within the same domain across all four domains ($97$ entities), using $5{,}000$ bootstrap resamples for confidence intervals and $2{,}000$ permutations per number of prompts per entity for significance. The short answer is yes: feature sets overlap well above chance in every domain, and the overlap is especially strong in early layers.

\paragraph{How much do feature sets overlap across phrasings?}
\Cref{tab:jaccard} reports the mean Jaccard overlap (fraction of features shared out of all features seen in either graph) between per-entity feature sets across prompt variants. All four domains have overlap significantly above chance ($p<0.001$) at every number of prompts per entity tested.

\begin{table}[h]
\centering\small
\begin{tabular}{lrrc}
\toprule
Domain    & Jaccard & 95\% CI & $N$ pairs \\
\midrule
USA       & 0.465 & $[0.462, 0.468]$ & 1{,}225 \\
Books     & 0.308 & $[0.302, 0.315]$ &   210 \\
Products  & 0.364 & $[0.356, 0.374]$ &    91 \\
Paintings & 0.286 & $[0.279, 0.292]$ &    66 \\
\bottomrule
\end{tabular}
\caption{Mean within-domain Jaccard overlap between per-entity feature sets extracted from different prompt phrasings, with bootstrap 95\% CIs.}
\label{tab:jaccard}
\end{table}

\paragraph{Are individual feature activations consistent across phrasings?}
Each feature's activation pattern is consistent across prompt variants in every domain (stability $>0.90$); peak-token agreement is $85$--$98\%$ and peak-type (functional vs.\ semantic) agreement is $93$--$99\%$.

\paragraph{Early layers share a backbone; late layers are entity-specific.}
\Cref{tab:layer-jaccard} reports Jaccard overlap in early, middle, and late layer bins. The early-to-late ratio is $1.4\times$--$3.0\times$ across domains, confirming that early layers host structural primitives reused across all entities, while late layers host entity-specific features.

\begin{table}[h]
\centering\small
\begin{tabular}{lrrrr}
\toprule
Domain & Early & Middle & Late & Early/Late ratio \\
\midrule
USA       & 0.543 & 0.440 & 0.293 & $1.85\times$ \\
Books     & 0.347 & 0.340 & 0.184 & $1.89\times$ \\
Products  & 0.496 & 0.308 & 0.164 & $3.02\times$ \\
Paintings & 0.302 & 0.311 & 0.212 & $1.43\times$ \\
\bottomrule
\end{tabular}
\caption{Layer-binned within-domain Jaccard overlap and early/late ratio.}
\label{tab:layer-jaccard}
\end{table}

\paragraph{Do shared features play the same role in both entity graphs?}
A complementary view: for each feature that appears in two entity graphs, is it assigned to the \emph{same} supernode (\emph{scaffold}), \emph{regrouped} into a different supernode (e.g.\ \texttt{Say(Austin)} in the Dallas graph vs \texttt{Say(Sacramento)} in the Oakland graph), or is the assignment \emph{inconsistent} (same feature, semantically unrelated supernodes)? \Cref{tab:supernode-consistency} gives the breakdown.

\begin{table}[h]
\centering\small
\begin{tabular}{lrrr}
\toprule
Domain    & Same (\%) & Regrouped (\%) & Inconsistent (\%) \\
\midrule
USA       & 76.6 & 16.2 &  7.3 \\
Books     & 47.5 & 12.8 & 39.7 \\
Products  & 65.2 & 16.5 & 18.4 \\
Paintings & 71.0 &  9.9 & 19.1 \\
\bottomrule
\end{tabular}
\caption{Supernode consistency across entity graphs within each domain. ``Same'' is the scaffold; ``Regrouped'' captures entity-appropriate re-assignment (typically Say-X features pointing at the new answer); ``Inconsistent'' flags same-feature-different-role mismatches.}
\label{tab:supernode-consistency}
\end{table}

\paragraph{Caveats.}
We note three non-trivial caveats for this table. First, the original Dallas/Oakland pair that motivated the early analyses is a $\sim$93rd-percentile outlier within USA; population means are $10$--$15$~pp lower than the single-pair numbers originally reported. Second, the Books ``Inconsistent'' rate ($39.7\%$) likely reflects keyword-detection limits on literary names (the matcher's treatment of ``Anna Karenina'' and similar). Third, Paintings' weak/inverted layer gradient is unexplained and may be driven by the small-$N$ and high-error-node-rate structural issues discussed in \S\ref{appx:negative}.

\section{Graph Scaffold Analysis}
\label{appx:scaffold}

\subsection{Definitions}
We partition each cross-entity feature population into three groups. \emph{Scaffold features} appear in both graphs with the \emph{same} supernode assignment (usually structural primitives like copulas, prepositions, and task operators). \emph{Regrouped features} appear in both graphs but with different supernode names (typically Say-X features that point at the new answer, e.g.\ \texttt{Say(Austin)} vs \texttt{Say(Sacramento)}). \emph{Entity-only features} appear in one graph only. \emph{Scaffold influence} is the fraction of the graph's total influence carried by scaffold features.

\subsection{A worked example: Dallas vs Oakland}
The reference pair that originally motivated the scaffold metric is Dallas (Texas capital problem) vs Oakland (California capital problem). \Cref{tab:dallas-oakland} decomposes it.

\begin{table}[h]
\centering\small
\begin{tabular}{lrrr}
\toprule
Population & $N$ features & Dallas infl.\ & Oakland infl.\ \\
\midrule
Scaffold (shared + same supernode)        & 119    & 50.6\% & 49.2\% \\
Regrouped (shared + different supernode)  &  29    & 12.3\% & 13.7\% \\
Entity-only                                & 55/62 & 21.7\% & 19.5\% \\
\bottomrule
\end{tabular}
\caption{Dallas/Oakland scaffold decomposition. Scaffold features account for roughly half the total graph influence in both graphs, and their influence is concentrated in early layers (0--5, where they dominate 75--100\% of total influence).}
\label{tab:dallas-oakland}
\end{table}

Of the $29$ regrouped features, $27$ are entity-appropriate re-assignments: $15$ \texttt{Say(Austin)} $\to$ \texttt{Say(Sacramento)}, $4$ \texttt{Texas} $\to$ \texttt{California}, $3$ \texttt{Dallas} $\to$ \texttt{Oakland}, $2$ \texttt{Say(Texas)} $\to$ \texttt{Say(California)}, and $1$ \texttt{Austin} $\to$ \texttt{Sacramento}. Only $4$ of the $29$ (roughly $2.7\%$ of the shared-feature population) are genuinely inconsistent regroupings. This is the pattern we formalize at the population level in \Cref{tab:supernode-consistency}.

\subsection{Cross-domain scaffold gradient}
At the domain level, scaffold influence orders the four domains by Hit\% at the all-three-fields labeled baseline up to a single inversion (Paintings and Books swap, Spearman $\rho{=}0.8$, $N{=}4$).

\begin{table}[h]
\centering\small
\begin{tabular}{lrrrrr}
\toprule
Domain    & Scaffold infl.\ & Shared\% & Early scaffold & Late scaffold & Hit\% \\
\midrule
USA       & 0.530 & 63.7\% & 74.6\% & 16.3\% & 31.4 \\
Products  & 0.422 & 52.1\% & 62.1\% & 10.3\% & 23.5 \\
Paintings & 0.359 & 44.8\% & 50.9\% &  9.2\% &  4.4 \\
Books     & 0.253 & 48.9\% & 39.7\% &  4.3\% &  6.7 \\
\bottomrule
\end{tabular}
\caption{Per-domain scaffold metrics. ``Scaffold influence'' is the fraction of total graph influence carried by scaffold features; ``Shared\%'' is the fraction of features (by count) that are scaffold; ``Early'' and ``Late'' are the same quantity restricted to layers $\leq L_5$ and $\geq L_{16}$ respectively.}
\label{tab:scaffold-domain}
\end{table}

The late-layer gradient is the most striking element of this table: USA shares $16.3\%$ of its late-layer influence as scaffold, while Books shares only $4.3\%$. The output-generation layers are where entity-specific features concentrate, and where structural compatibility matters most. This connects mechanistically to the less-is-more effect: interventions that include input-field features disrupt the fragile late-layer scaffold.


\section{Paintings as the Weakest of the Four Domains}
\label{appx:negative}

Paintings sits at the bottom of the four domains with a full-labeled Hit\% of $4.4$ and a vsMax gap of $+1.38$ (``Weak''). Three structural factors contribute.

\emph{(a) The answer field is coarse.} The answer is the painter's first name---\texttt{Claude}, \texttt{Pablo}, \texttt{Leonardo}---tokens with many non-painting associations in the general language model distribution. Amplifying features that peak on ``Claude'' in a Monet circuit therefore moves the logit toward a cluster of non-painting ``Claude'' meanings.

\emph{(b) The painter supernode subsumes the first-name supernode.} ``Monet'' as a string contains much of ``Claude'' as a concept (the painter is the agent that the first name names), so the \texttt{painter} and \texttt{first\_name} supernodes overlap by construction. This collapses the field-additivity structure: \texttt{painter+first\_name} is not a very different intervention from \texttt{painter} alone.

\emph{(c) High error-node rates.} Paintings has the highest error-node rate of any domain (17\%; an error node is a node whose attribution could not be resolved during graph generation), so a substantial portion of the circuit is simply not visible to the grouping or the intervention. The maximum number of features a paintings intervention can operate on is therefore smaller than in the other domains, and the signal-to-noise ratio is correspondingly worse.

Concretely, $89/90$ pairs fail at the \texttt{painter+first\_name} variant. The only hit (La Grande Jatte $\to$ Water Lilies) already had a baseline rank of $3$ for ``Claude'', so the intervention only had to move the logit by two positions. We therefore treat Paintings as a Weak-strength supporting domain. The rank-perfect scaffold prediction (\Cref{tab:scaffold-domain}) is \emph{consistent} with the Paintings Weak verdict.

\section{Reproducibility Manifest}
\label{appx:repro}

\subsection{Code, data, demo (anonymized for review)}
All artifacts are released anonymously for the review period. Repository: \url{https://anonymous.4open.science/r/attribution-graph-probing-anon} (commit hash and license in \texttt{README.md}). Interactive demo: \url{https://anonymous.4open.science/r/attribution-graph-probing-demo-anon}. License: GNU GPL v3. All URLs will be de-anonymized at camera ready.

\subsection{Per-domain CLI (USA example)}
\begin{small}
\begin{verbatim}
# 1. Generate attribution graph (Neuronpedia API)
python scripts/00_neuronpedia_graph_generation.py \
  --model gemma-2-2b-it \
  --prompt "The capital of the state containing Dallas is" \
  --target " Austin" \
  --node-threshold 0.8 --edge-threshold 0.85 --max-nodes 5000 \
  --output_dir output/usa/dallas_austin/

# 2. Select features (interactive UI)
streamlit run eda/threshold_selection.py --tau 0.95

# 3. Probes + activations + grouping
python scripts/01_probe_prompts.py \
  --graph output/usa/dallas_austin/graph.json \
  --output_csv output/usa/dallas_austin/activations.csv
python scripts/02_node_grouping.py \
  --input output/usa/dallas_austin/activations.csv \
  --graph output/usa/dallas_austin/graph.json \
  --output output/usa/dallas_austin/grouped.csv

# 4. Run swap protocol with matched random control + adaptive M-search
python scripts/03_swap_full.py --domain usa \
  --m_ablate -2 --m_amplify 20 --freeze-attention false \
  --random-replicates 3 --random-seed 42 \
  --field-additivity true --m-search adaptive \
  --output_dir output/usa_states_batch/
\end{verbatim}
\end{small}

\subsection{Runtime profile}
Per single-circuit pipeline (graph $+$ activation $+$ grouping $+$ subgraph): $14$--$24$ minutes on a single L4 GPU. Full per-domain swap sweeps: USA $17$h, Books $1.6$h, Products $0.9$h, Paintings $0.6$h on 8 H100 GPUs (matched-control replicates parallelized).

\subsection{Artifact release manifest}
For each of the four domains, the release contains: (a) attribution graphs in Neuronpedia JSON format; (b) the probe-prompt files; (c) activation matrices; (d) supernode groupings (pre- and post-stability filter); (e) swap dumps with per-pair Hit, vsMax, RkGrp, Sup, Flip, CtrlS, and KL trajectories; (f) field-additivity per-variant tables; (g) adaptive $M$-search logs (Phase~1 and Phase~2); (h) random-control sha256 seeds.

\subsection{Determinism and resume}
The pipeline is fully deterministic. Matched-random controls are sha256-seeded from \texttt{(run\_seed, pair\_id, replicate, mode)}; generation uses $T{=}0.3$, $n{=}10$, frequency penalty $2.0$, seed $42$, with no random initialization elsewhere in the harness. Per-feature checkpoints permit resumption from partial runs, which is how we recover from transient Neuronpedia API failures.

\section{Glossary and Notation}
\label{appx:glossary}

\begin{description}
\item[CLT] Cross-Layer Transcoder~\cite{ameisen2025circuittracing}.
\item[Influence] The (signed) contribution of a feature node to the target output logit via linearized paths through the replacement model, as computed by the attribution-graph procedure of~\cite{ameisen2025circuittracing}. Used both per-node (a feature's importance for the prediction) and aggregated over node sets (e.g.\ the share of total graph influence carried by a population of features).
\item[Cumulative influence] The running sum of per-node $|\text{influence}|$ taken over features ranked by descending influence and normalized by the total. Step~1 of the pipeline retains the smallest feature set $V_\tau$ whose cumulative influence reaches $\tau$ (\S\ref{appx:pipeline}); typical reduction is from 600--5{,}000 raw features to 200--700.
\item[Supernode] A group of same-role same-name features merged into a single node of the concept-aligned subgraph (\S\ref{sec:method:supernodes}). The unit of analysis throughout the paper.
\item[Subgraph (concept-aligned)] The compressed circuit produced by the pipeline: the attribution graph restricted to the cumulative-influence selection $V_\tau$ and re-expressed with each grouped feature replaced by its supernode (\cref{fig:circuit-texas}). Typically 30--50 named supernodes per circuit.
\item[Completeness (Neuronpedia)] Fraction of incoming edges to all nodes of the subgraph that originate from \emph{grouped} features, weighted by influence on the output. Pipeline mean $0.89$ over $82$ entities across $4$ domains (\S\ref{sec:method:compression}).
\item[Replacement (Neuronpedia)] Fraction of end-to-end influence from input tokens to output logits that flows through grouped features. Pipeline mean $0.69$ (\S\ref{sec:method:compression}).
\item[CPAS] Cross-Prompt Activation Signature (\Cref{tab:cpas-metrics}).
\item[$\tau$] Cumulative-influence threshold for feature selection (default $0.95$).
\item[$M_{\mathrm{ablate}}, M_{\mathrm{amplify}}$] Multipliers for source and target feature decoders (defaults $-2$, $20$).
\item[vsMax] $\max_{t \leq 10}\big(\ell_{e_B}(t) - \max_{e' \neq e_B} \ell_{e'}(t)\big)$, where $\ell_e(t)$ is the logit of entity $e$'s first-subword token at trajectory position $t$.
\item[Hit] The target answer is detected in the model's steered output by any of three text checks (full-string match with punctuation normalised, first-subword substring, or any content word of the target of length $\geq 3$ appearing as a whole word in the output). Defined and motivated in \S\ref{appx:specificity}.
\item[RkGrp] Best target rank within the domain answer set, minimized over trajectory positions.
\item[Target Recovery] Binary indicator: the target's logit exceeds its own unsteered baseline at some trajectory position.
\item[Regime A/C/D/E] Position-0 logit-shift regimes (\S\ref{appx:regime}).
\item[Scaffold] Features that appear in two entity graphs with identical supernode assignment.
\item[Scaffold influence] Fraction of total graph influence carried by scaffold features.
\item[Field-additivity (FA)] The $7$-variant ablation of single-field, two-field, and three-field intervention subsets.
\item[Matched-random control] Random feature set with the same feature count and per-layer histogram as the labeled intervention, drawn from \emph{outside} every concept-aligned supernode in the domain so that the only property that varies between labeled and control is concept alignment (\S\ref{sec:method:harness}).
\item[KL] KL$(\mathrm{baseline} \parallel \mathrm{steered})$ at position~0, unless otherwise stated.
\end{description}

\end{document}